\documentclass{article}

\usepackage{arxiv}
\usepackage{authblk}
\usepackage[utf8]{inputenc} 
\usepackage{hyperref}    
\usepackage{doi}
\usepackage{titling}
\usepackage{amsfonts}
\usepackage{graphicx}
\usepackage{epstopdf}
\usepackage{mathtools}
\usepackage{color}
\usepackage{amsmath}  
\usepackage{cases}
\usepackage{amssymb}
\usepackage{booktabs}
\usepackage{multirow}
\usepackage{amsmath}  
\usepackage{algorithm,algorithmicx,algpseudocode}
\usepackage{amssymb}
\usepackage{url}
\usepackage{multicol}
\usepackage{cite}
\usepackage{enumitem} 
\usepackage{booktabs}
\usepackage{varwidth}
\newsavebox\tmpbox
\usepackage{soul}
\usepackage[caption=false]{subfig}
\usepackage{csquotes}

\newcommand{\norm}[1]{\left\lVert#1\right\rVert}
\renewcommand{\shorttitle}{Patch-Based Image Restoration using Expectation Propagation}
\def\middlebreak {\nulldelimiterspace0pt
\right.\allowbreak\mskip 0mu plus .5mu \nulldelimiterspace0pt\left.}

\algdef{SE}[SUBALG]{Indent}{EndIndent}{}{\algorithmicend\ }%
\algtext*{Indent}
\algtext*{EndIndent}

\hypersetup{
pdftitle={A template for the arxiv style},
pdfsubject={q-bio.NC, q-bio.QM},
pdfauthor={David S.~Hippocampus, Elias D.~Striatum},
pdfkeywords={First keyword, Second keyword, More},
}

\title{Patch-Based Image Restoration using Expectation Propagation}

\author{Dan Yao\thanks{Email: \texttt{dy14@hw.ac.uk}}, ~Stephen McLaughlin\thanks{Email: \texttt{S.McLaughlin@hw.ac.uk}}, ~Yoann Altmann\thanks{Email: \texttt{Y.Altmann@hw.ac.uk}}\\ 
	School of Engineering and Physical Sciences\\
	Heriot-Watt University\\
	Edinburgh, EH14 4AS, United Kingdom \\
}

\begin{document}
\maketitle

\begin{abstract}
This paper presents a new Expectation Propagation (EP) framework for image restoration using patch-based prior distributions. While Monte Carlo techniques are classically used to sample from intractable posterior distributions, they can suffer from scalability issues in high-dimensional inference problems such as image restoration. To address this issue, EP is used here to approximate the posterior distributions using products of multivariate Gaussian densities. Moreover, imposing structural constraints on the covariance matrices of these densities allows for greater scalability and distributed computation. While the method is naturally suited to handle additive Gaussian observation noise, it can also be extended to non-Gaussian noise. Experiments conducted for denoising, inpainting and deconvolution problems with Gaussian and Poisson noise illustrate the potential benefits of such a flexible approximate Bayesian method for uncertainty quantification in imaging problems, at a reduced computational cost compared to sampling techniques.  
\end{abstract}



\section{Introduction}
\subsection{Problem Formulation}

In this paper, we address the problem of image restoration which consists of estimating an unknown image from its degraded observation, e.g., noisy, blurry, or missing pixels.  The (vectorized) observed image $\pmb y \in \mathbb{R}^{N}$ is modeled as a known linear transformation of the unknown image of interest $\pmb x\in \mathbb{R}^{N}$. The linear transformation is denoted by $\textbf{H}{\pmb x}$ and the matrix ${\mathbf H} \in {\mathbb R}^{N\times N}$ is either diagonal (e.g., as in denoising and inpainting problems) or diagonalizable in a transformed domain (e.g., as in deconvolution problems). The observation noise can be either Gaussian, independently and identically distributed (i.i.d.) or more complex, e.g., non-Gaussian. Adopting a Bayesian approach, the observation model described by the likelihood function $f_{y|x}(\pmb y|{\bf H}\pmb x)$, and a prior distribution $f_x(\pmb x|\pmb \theta)$ parameterized by $\pmb \theta$ and based on image patches, are adopted to perform image restoration. Following Bayes' theorem, the posterior distribution of $\pmb x$ is
\begin{equation}
    f(\pmb x|\pmb y,\pmb \theta) =\frac{ f_{y|x}(\pmb y|{\bf H}\pmb x)f_x(\pmb x|\pmb \theta)}{\displaystyle\int f_{y|x}(\pmb y|{\bf H}\pmb x)f_x(\pmb x|\pmb \theta){\rm d}{\pmb x}}.
\label{Eq: exact posterior distribution}
\end{equation}

Exact inference beyond Maximum A Posteriori (MAP) estimation using \eqref{Eq: exact posterior distribution} is however extremely difficult for high-dimensional imaging problems, due to the high-dimensional, usually intractable integral, in the denominator. Prior to describing the proposed approximate Bayesian method enabling scalable inference, we briefly describe related approaches previously investigated for Bayesian image restoration.

\subsection{Related Work}
{\textbf {Designing the prior model.}} 
The choice of suitable prior models depends on the amount of information available about the images expected to be recovered and generally relies on a trade-off between the quality of the resulting images and the computational complexity of the algorithm(s) required to perform Bayesian inference. Traditional image priors rely on explicit distributions modeling statistical properties of images in some transform domain \cite{figueiredo2003algorithm,abramovich1998wavelet,bioucas2006bayesian}, or Markov random fields (MRFs) \cite{rue2005gaussian}. Such prior models are often generic, i.e., they do not encode strong prior information and have been successfully applied to a variety of imaging problems, e.g., medical imaging and astroimaging, for the restoration of images corrupted by different noise models. 

Over the last few years, patch-based image restoration methods have also received a lot of interest. Such approaches are particularly attractive from a computational point of view as learning patch-based priors and using them to restore patches/images can be simpler than learning and using global image priors \cite{zoran2011learning, niknejad2019external}. In practice, overlapping patches are often found to be crucial to achieve performance close to state-of-the-art image restoration methods\cite{elad2010sparse}. There are two main categories of methods in literature to learn/build patch-based priors. In the first category, patch-based prior models are learned on image patches which are assumed to be independent. Such methods include for example \cite{teodoro2015single,niknejad2015image,teodoro2020block,zoran2011learning,zhang2016gaussian} using Gaussian Mixture Models (GMMs) as a patch-based prior. Conversely, the second category of methods accounts for dependencies between overlapping patches while learning a patch-based prior. Fields-of-Experts (FoE) priors \cite{roth2005fields} are typical examples of such methods. However, training FoE models is computationally intensive compared to the first category of methods. In this paper, to keep the overall inference process (prior design and image restoration) computationally tractable, we consider expressive GMM models built using tools from the first category of methods, as priors of image patches. To alleviate the usual blocky artifacts when restoring images from non-overlapping patch estimates, here we use GMM priors for different partitions of the image of interest.  Individual image posteriors are obtained from the different partitions and they are combined to obtain the final image estimates (approximate mean and marginal variances).

Recently, more advanced priors, such as plug-and-play priors \cite{venkatakrishnan2013plug,romano2017little} and  priors trained by a deep neural network \cite{ulyanov2018deep,mataev2019deepred,vaksman2020lidia,zhang2017beyond}, have become popular in image restoration methods and shown leading performance in many image restoration tasks. However, most of these priors are limited to delivering point estimates, without clear rules to quantify the resulting uncertainties. The GMM patch-based prior considered in this work helps to built a flexible and scalable inference framework to provide both point estimates and pixel-wise posterior uncertainty measures. Since this prior models non-overlapping image patches independently, it allows for greater scalability in solving high-dimensional imaging problems.
Being a conjugate prior for Gaussian observation model \cite{bernardo2009bayesian}, it is extended in this work to solve more complex image restoration problems beyond denoising with Gaussian noise, such as observations corrupted by Poisson noise. 

{\textbf {Choosing appropriate point estimators and summary statistics}}. 
The majority of Bayesian methods using patch-based image priors adopt the MAP estimator as the preferred point estimator of $\pmb x$ as it can be computed using powerful optimization tools.
In the case of patch-based priors e.g., \cite{niknejad2015image,teodoro2020block,zoran2011learning,zhang2016gaussian,roth2005fields}, the resulting posterior distribution may not be convex nor unimodal. Nonetheless, convex optimization tools can be used to optimize (local) bounds of the log-posterior distribution \cite{niknejad2015image}.
An alternative point estimator is the Minimum Mean Squared Error (MMSE) estimator, which, when combined with associated uncertainty measurement, is particularly useful in problems where the posterior distributions are skewed. This is typically the case in imaging situations where the observations are corrupted by Poisson-like, multiplicative noise. Unfortunately, exact MMSE inference for high-dimensional image restoration problems, is often intractable, or computationally expensive. In the context of patch-based priors, while MMSE estimation is not generally possible at the image level, it is however possible to compute MMSE patch estimates, and to combine these estimates by averaging. The MMSE estimates can be either obtained analytically \cite{teodoro2015single} or approximated by Monte Carlo sampling \cite{niknejad2019external}. These approaches, which have been proposed for image denoising and inpainting, are closely related to our work. However, we extend their capabilities to image deconvolution, i.e., when degradation operator ${\mathbf H}$ is non-diagonal. 

While most MAP-based methods can produce satisfactory point estimates for restored images, the algorithms are generally unable to directly quantify the uncertainty in the solutions delivered \cite{pereyra2017maximum}. For log-concave posteriors however, MAP estimation methods based on convex optimization can identify regions of high-posterior densities, that can be used to hypothesis testing \cite{repetti2019scalable}. Although numerous patch-based image restoration methods have been proposed (e.g.,  \cite{teodoro2015single,teodoro2020block,roth2005fields,zoran2011learning,niknejad2019external,niknejad2015image}), such priors are rarely studied beyond MAP estimation. This is probably because several challenges arise when considering MMSE estimation and uncertainty quantification at the image level using priors based on overlapping patches. 

{\textbf {Approximate estimation.}} 
When exact computation of posteriori summary statistics is not possible, approximation strategies have been proposed to approximate the intractable posterior distribution. These strategies can be stochastic, deterministic or hybrid. Stochastic methods are carried out by sampling, such as importance sampling \cite{niknejad2019external} or Markov chain Monte Carlo (MCMC) sampling. However, sampling methods for accurate approximation of high-dimensional integrals are still computationally expensive and not yet fully scalable for fast inference. While estimating posterior means might only require relatively short Markov chains, estimating higher-order moments is much more complicated, especially when the samplers used suffer from poor mixing properties as the dimension of the problem increases. Recent advances in high-dimensional samplers for non-smooth posterior distributions, e.g., using proximal MCMC methods \cite{durmus2018efficient} have led to significant speed-up, at the cost of a tunable accuracy loss induced by model augmentation and/or posterior smoothing. 
Variational inference represents another family of approximate methods which complement sampling methods \cite{bishop2006pattern, Blei_2017}. Their basic principle is to approximate the target distribution $f(\pmb x|\pmb y,\pmb \theta)$ by a more tractable distribution whose moments are easier to compute. This family includes Variational Bayes (VB) methods and methods relying on Expectation Propagation (EP) \cite{minka2013expectation, opper2000gaussian,gelman2014expectation}, which has recently received growing interest for high-dimensional problems. The method proposed in this work relies on EP, whose motivations and imaging applications will be further reviewed in the next paragraph.

{\textbf{Expectation Propagation for imaging problems.}} 
EP algorithms have been successfully applied to imaging  problems with different observation models and different image priors. For example, in \cite{seeger2011fast}, a Laplace prior enforcing sparsity of the wavelet coefficients or of the image gradient was adopted to address image deconvolution problems in the presence of Gaussian noise. Gaussian processes and sparse linear models \cite{seeger2007bayesian,seeger2011large} were considered as prior distributions when combined with a Poisson observation model. 
Despite the success of EP applying to (generalized) linear regression models \cite{hernandez2015expectation,altmann2019expectation,kim2018expectation,heess2013learning} and imaging restoration problems  \cite{braunstein2020compressed,Muntoni_2019,braunstein2017analytic,2019InvPr}, to the best of our knowledge, EP has not yet been combined with patch-based priors. 

Identifying uncertainties and posterior correlation in high-dimensional image restoration problems usually requires the computation of large covariance/precision matrices. This computational cost is presented when performing exact inference but also when using approximate methods like VB/EP.  Two main approximate covariance matrix types have been exploited in EP methods to model or not the correlation between variables. Using full covariance matrices requires either the inversion of large matrices or a large number of sequential updates leveraging rank-1 optimization of large matrices \cite{seeger2007bayesian, cunningham2011gaussian, 2019InvPr}. To reduce computational time and memory requirements, diagonal covariance matrices are often used within EP. This effectively neglects local correlation between groups of variables. However, it has been shown that if significant correlation is omitted, EP can produce poor results \cite{potetz2013whitened}.

This paper considers block-diagonal covariance matrices to capture partial correlation between variables. Instead of considering a vanilla EP factorization and implementation resulting in a large number of sequential updates, the proposed EP method consists of only two or three factors, depending on the noise model considered. This leads to EP algorithms with only two to three sequential updates at each iteration, each update being highly parallelizable and allowing distributed inference. This is a strategy similar to the factorization scheme used in AMP/GAMP/VAMP \cite{donoho2009message, rangan2011generalized, schniter2016vector}, which enables faster inference. However, in contrast to AMP/GAMP/VAMP, not all the covariance matrices are constrained to be isotropic nor diagonal in the proposed EP method.

The goal of this work is to propose a new framework of EP algorithms with existing patch-based priors, enabling approximate uncertainty quantification and visualization for imaging problems. The main contributions of this paper can be summarized as follows.

\subsection{Contributions}
\begin{enumerate}
    \item [1.] A new EP algorithm with GMM patch-based prior is, to the best of our knowledge, proposed for the first time for image restoration. Approximate MMSE estimates and associated posterior uncertainties are directly obtained from the moments of the Gaussian approximation of the exact posterior distribution. The algorithm can be applied to large images thanks to the structure imposed to the covariance matrices of the approximating distributions. 
    \item [2.] A variational Expectation Maximization (EM) algorithm is also used to estimate GMM hyperparameters using the EP approximation of the exact posterior distribution. This allows the prior model to be adjusted for each image and the resulting algorithm does not require significant user supervision.
    \item [3.] While the method is naturally suited to handle Gaussian noise models, it is extended for non-Gaussian noise cases, and the Poisson noise model is used as illustration. 
\end{enumerate}

The reminder of this paper is organized as follows. A new patch-based expectation propagation algorithm for Gaussian observation models is proposed in Section \ref{Sec: patch-basedEP_Gaussian_observation}. In Section \ref{Sec: Patch-based EP with non-Gaussian noise}, the algorithm is generalized to non-Gaussian observation models, using Poisson noise as a running example. Section \ref{Sec: Final Estimation using Geometric Average} describes the final image restoration strategy relying on model averaging. In Section \ref{Sec: EP-EM Strategy for Hyperparameter Estimation}, an EP-EM strategy is proposed to estimate hyperparameters of the GMM priors. Experimental results are presented in Section \ref{Sec: Experimental Results}. Finally, conclusions and discussions are reported in Section \ref{Sec: Discussion and Conclusions}. 

\section{Patch-based EP algorithm with Gaussian noise}
\label{Sec: patch-basedEP_Gaussian_observation}

\subsection{Exact Bayesian model with GMM prior}

Given a vectorized image $\pmb x$ composed of $\sqrt{N} \times \sqrt{N}$ pixels, we consider a set of $J$ small non-overlapping patches $\{\pmb x_j\}_{j=1,\dots,J}$ of size $\sqrt{r} \times \sqrt{r}$. Each pixel in the image is assigned to a single patch, leading to $J=N/r$. The patches $\{\pmb x_j\}_{j=1,\dots,J}$ are assigned independently the same GMM prior model 
\begin{equation}
    \pmb x_j|\pmb \theta \sim \sum\limits_{k=1}^K \omega_k \mathcal{N}(\pmb x_j;\tilde{\pmb \mu}_k, \widetilde{\bf C}_k),
\label{Eq: x_j_GMM}    
\end{equation}
\noindent where $K$ is the number of mixture components, $\omega_k$, $\tilde{\pmb \mu}_k$, $\widetilde{\bf C}_k$ are respectively the weight, mean, and covariance matrix of the $k$-th Gaussian component, and $\boldsymbol{\theta}=\{\omega_k, \tilde{\pmb \mu}_k, \widetilde{\bf C}_k\}_{k=1,\ldots,K}$. Moreover, the positive weights $\{\omega_k\}_{k=1,\dots,K}$ sum to one. 
In this section and Sections \ref{Sec: Patch-based EP with non-Gaussian noise} and \ref{Sec: Final Estimation using Geometric Average}, the GMM parameters in $\pmb \theta$ are assumed to be known. Possible procedures to learn these parameters from training images will be discussed in Section \ref{Sec: EP-EM Strategy for Hyperparameter Estimation}, together with our proposed method to refine them. The global image prior model $f_x(\pmb x|\pmb \theta)$ can be expressed as
\begin{equation}
    f_x(\pmb x|\pmb \theta) = \displaystyle\prod\limits_{j=1}^J  \sum\limits_{k=1}^K \omega_k \mathcal{N}( \pmb x_j;\tilde{\pmb \mu}_k, \widetilde{\bf C}_k).
\label{Eq: proposed GMM patch-based prior} 
\end{equation}

In this section, we assume that the observations $\pmb y$ are corrupted by i.i.d. Gaussian noise with known noise variance $\sigma^2$. Thus, the likelihood function can be expressed as 
\begin{eqnarray}
f_{y|x} (\pmb y|{\bf H}\pmb x) = \prod\limits_{n=1}^N {\mathcal N}(y_n;\pmb h_n \pmb x,\sigma^2),
\end{eqnarray}
where $\{\pmb h_n\}_{n=1,\dots,N} \in {\mathbb R}^{1\times N}$ are the row vectors of $\bf H$. The resulting exact posterior distribution of $\pmb x$ is given, up to a multiplicative constant, by
\begin{equation}
f(\pmb x|\pmb y,\pmb \theta) \propto \left[\prod\limits _{n=1}^N {\mathcal N}(y_n;\pmb h_n\pmb x,\sigma^2)\right]f_x(\pmb x|\pmb \theta).
\label{Eq: exact_posterior_Gaussian}
\end{equation}

The goal of the proposed EP algorithm is to find a Gaussian distribution $Q(\pmb x)$ that closely approximates the exact posterior $f(\pmb x|\pmb y,\pmb \theta)$, in particular when computing the MMSE estimate $ {\mathbb E}_{f(\pmb x|\pmb y,\pmb \theta)}[\pmb x]$ is intractable. Using the tractable approximating distribution $Q(\pmb x)$, the exact MMSE estimate can be approximated by ${\mathbb E}_{f(\pmb x|\pmb y,\pmb \theta)}[\pmb x] \approx {\mathbb E}_{Q(\pmb x)}[\pmb x]$ and the posterior covariance matrix can be approximated by the covariance matrix of $Q(\pmb x)$. 

Note that in cases where ${\bf H}^T{\bf H}$ is diagonal, e.g., in denoising or inpainting problems, the EP algorithm discussed in this section is not required to perform a posteriori inference. Indeed, as discussed in \cite{teodoro2015single}, in that case, the posterior distribution in \eqref{Eq: exact_posterior_Gaussian} is tractable and reduces to patch-wise independent GMMs. However, we still discuss this case as a starting point of this paper as the resulting EP updates can be used for more complex problems where the noise is non-Gaussian, as in Section \ref{Sec: Patch-based EP with non-Gaussian noise}.

\subsection{EP approximation strategy} 
As alluded to above, EP methods rely on approximating a distribution by a product of (unnormalized) probability density function  (p.d.f.) or probability mass function  (p.m.f.) from the exponential family. The factors of that product are then updated sequentially. To alleviate a large number of sequential updates, we consider an  approximation of $f(\pmb x|\pmb y,\pmb \theta)$ with two factors. The first factor, which relates to $f_x(\pmb x|\pmb \theta)$ (seen as a function of $\pmb x$) and which depends implicitly on $\boldsymbol{\theta}$, is denoted by $q_{x,0}(\pmb x)$. The second factor, which relates to 
$f_{y|x}(\pmb y|{\mathbf H}\pmb x)$ (seen as a function of $\pmb x$) and which implicitly depends on $\pmb y$, is denoted by $q_{x,1}(\pmb x)$. These two factors are, for simplicity, unnormalised multivariate Gaussian densities whose parameters are defined as 
\begin{equation}
q_{x, 0}(\pmb x)  \propto {\mathcal N}(\pmb x;\pmb m_{x,0},{\bf \Sigma}_{x, 0})\quad \textrm{and}\quad q_{x, 1}(\pmb x) \propto {\mathcal N}(\pmb x;\pmb m_{x,1},{\bf \Sigma}_{x, 1}).
\end{equation}

\begin{figure}[!ht]
\centering
\includegraphics[width=0.5\textwidth]{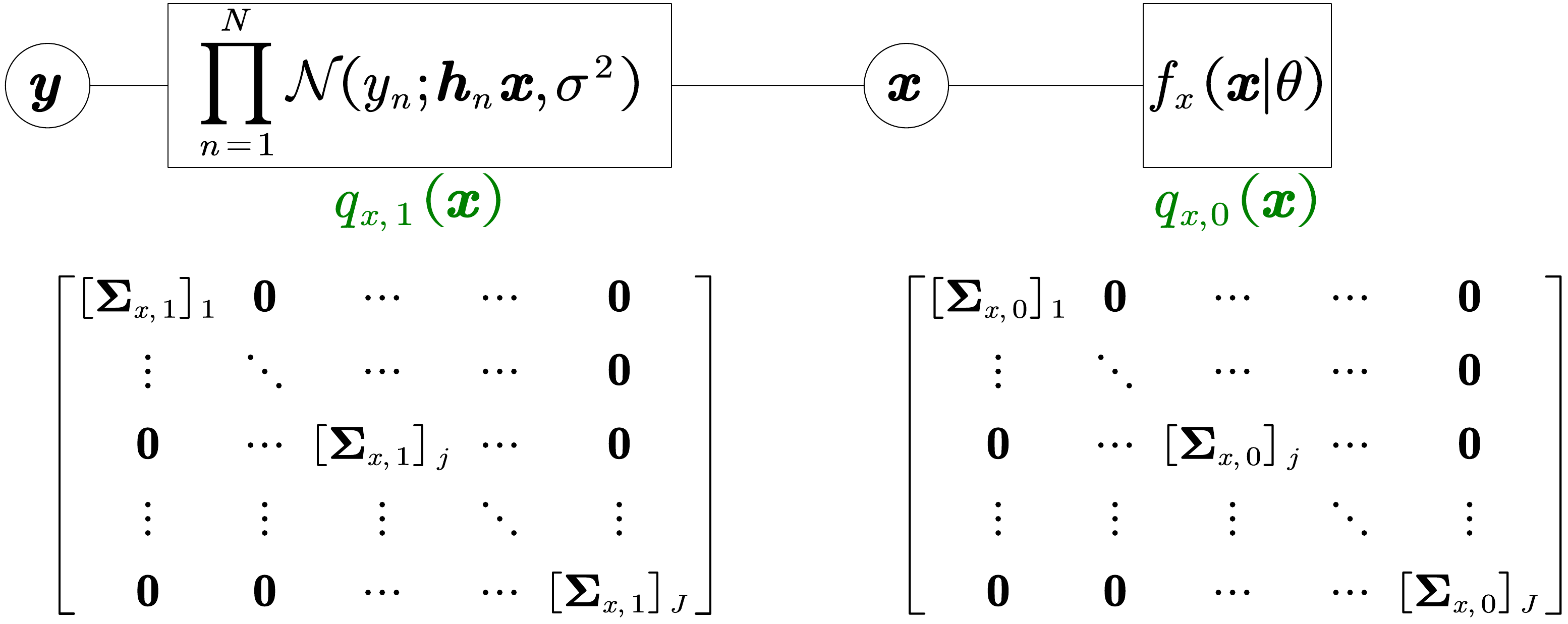}
\caption{Factor graph used to perform EP approximation for Gaussian likelihood and GMM patch-based prior. The rectangular boxes (resp. circles) represent the factors (resp. variables) involved in the factorized exact posterior distribution. The corresponding EP approximating factors are shown in green.}
\label{fig: factor_graph_1}
\end{figure}

A factor graph representing this EP factorization is depicted in Figure \ref{fig: factor_graph_1}. The resulting approximation of $f(\pmb x|\pmb y, \pmb \theta)$ is given by $Q(\pmb x) \propto q_{x,1}(\pmb x)q_{x,0}(\pmb x)$. To find the variational parameters of $Q(\pmb x)$, EP reduces to solving iteratively the two following KL divergence minimization problems  
\begin{subnumcases}{\label{Eq: KL minimization for approx Gaussian likeli}}
& $\mathop{\min}\limits_{q_{x, 0}(\pmb x)} KL\left(f_{x}(\pmb x|\pmb \theta)q_{x, 1}(\pmb x)||q_{x, 0}(\pmb x)q_{x, 1}(\pmb x)\right)$, \label{Eq: q_0_x_Gauss}
   \\
& $\mathop{\min}\limits_{q_{x, 1}(\pmb x)}KL\left(\prod\limits _{n=1}^N {\mathcal N}(y_n;\pmb h_n\pmb x,\sigma^2)q_{x, 0}(\pmb x)||q_{x, 1}(\pmb x)q_{x, 0}(\pmb x)\right)$. \label{Eq: q_1_x_Gauss}
\end{subnumcases}

Traditionally, the densities obtained by removing a single factor from $Q(\pmb x)$, i.e., $q_{x, \backslash i}(\pmb x) \propto Q(\pmb x)/q_{x, i}(\pmb x)$ ($\forall i \in \{0,1\}$), are referred to as cavity distributions. Here, we simply have $q_{x, \backslash 0}(\pmb x)=q_{x,1}(\pmb x)$ and $q_{x, \backslash 1}(\pmb x)=q_{x,0}(\pmb x)$. Moreover, the (unnormalized) densities obtained by multiplying these cavity distributions by their corresponding true factors are referred to as tilted distributions. Here, we have two tilted distributions, $P_0(\pmb x)\propto f_{x}(\pmb x|\pmb \theta)q_{x, 1}(\pmb x)$  and $P_1(\pmb x)\propto \prod\limits _{n=1}^N {\mathcal N}(y_n;\pmb h_n\pmb x,\sigma^2)q_{x, 0}(\pmb x)$, each being associated with one of the KL divergence minimization problems in \eqref{Eq: KL minimization for approx Gaussian likeli}.

Within the traditional EP framework where the problems in \eqref{Eq: KL minimization for approx Gaussian likeli} are solved without additional constraints on the covariance matrices ${\bf \Sigma}_{x, 0}$ and ${\bf \Sigma}_{x, 1}$, solving the KL divergence minimization problems reduces to (\romannumeral 1) computing the mean and covariance matrix of each tilted distribution, denoted by ${\mathbb E}_{P_i}[\pmb x]$ and ${\rm Cov}_{P_i}(\pmb x)$ (for $i\in\{0,1\}$), and (\romannumeral 2) matching those moments with those of $Q(\pmb x)$, denoted by $\pmb m_*$ and ${\bf \Sigma}_*$. Since $Q(\pmb x)$ is the product of two Gaussian densities, these approximate moments are given by
\begin{equation}
{\bf \Sigma}_*^{-1} =  {\bf \Sigma}_{x, 0}^{-1} + {\bf \Sigma}_{x, 1}^{-1},\quad 
\pmb m_*  =  {\bf \Sigma}_*\left( {\bf \Sigma}_{x, 0}^{-1} \pmb m_{x,0}+ {\bf \Sigma}_{x, 1}^{-1}\pmb m_{x,1}\right).
\end{equation}
At each EP iteration, the algorithm produces, sequentially, new values of $(\pmb m_{x,0}, {\bf \Sigma}_{x,0})$ and $(\pmb m_{x,1}, {\bf \Sigma}_{x,1})$, that minimize the first and second line of \eqref{Eq: KL minimization for approx Gaussian likeli}, respectively. However, this simple strategy often does not apply directly in the multivariate case as, for instance, the covariance matrix ${\mathbf \Sigma}_{x,1}^{-1} = \left({\rm Cov}_{P_1}(\pmb x)\right)^{-1}-{\bf \Sigma}_{x, 0}^{-1}$ is not ensured to be positive definite. To alleviate this issue, the covariance matrices ${\bf \Sigma}_{x, 0}$ and ${\bf \Sigma}_{x, 1}$ are often constrained to be diagonal or isotropic. This drastically simplifies the resolution of the problems in \eqref{Eq: KL minimization for approx Gaussian likeli} when subject to the symmetry and positive-definiteness constraints applied to the covariance matrices. However, while imposing such diagonal structural constraints on ${\bf \Sigma}_{x, 0}$ and ${\bf \Sigma}_{x, 1}$ can stabilize the EP algorithm, it constrains $Q(\pmb x)$ to neglect posterior correlations between the elements of $\pmb x$ and thus potentially degrades the quality of the approximation (as will be further discussed in Section \ref{Sec: Experimental Results}). In this work, we propose to use less restrictive constraints and allow ${\bf \Sigma}_{x, 0}$ and ${\bf \Sigma}_{x, 1}$ to be block-diagonal matrices, as discussed next.

\subsection{Block-diagonal covariance matrix approximation}
Spatial correlation between pixels is induced by the GMM priors (a priori known correlation), but it can also be induced by the likelihood factor, e.g., in deconvolution problems. While estimating and handling full, unstructured covariance matrices is not tractable for large images due to computationally intensive matrix inversions, it is possible to partially capture posterior correlation by imposing structural constraints on $({\bf \Sigma}_{x, 1}, {\bf \Sigma}_{x, 0})$, and thus on ${\bf \Sigma}_*$. Enforcing block-diagonal structures appears to be a natural and computationally attractive option. Indeed, the prior model in \eqref{Eq: proposed GMM patch-based prior} induces, by construction, spatial correlation within image patches and a block-diagonal structure in the prior covariance matrix of $\pmb x$ (provided that the pixels are sorted properly in $\pmb x$). Moreover, the degradation operator $\textbf{H}$ formed by convolution kernels are diagonally dominant. Thus, we propose to use the same block-diagonal structure for ${\bf \Sigma}_{x, 1}$ and ${\bf \Sigma}_{x, 0}$, which is the $r \times r$ block structure induced by the patch-based prior. For any patch-based image partition, it is possible to use this block-diagonal structure by re-ordering the observations in $\pmb y$, the pixels in $\pmb x$ and the rows of $\mathbf H$.

\subsection{EP update schemes with block-diagonal covariance matrices}
\label{subsec: EP update for block-diagonal covariance with positive definite constriant}
We can now discuss how the problems in \eqref{Eq: KL minimization for approx Gaussian likeli} can be solved efficiently assuming that ${\bf \Sigma}_{x, 1}$ and ${\bf \Sigma}_{x, 0}$ share the same block-diagonal structure. The simplest is the update in \eqref{Eq: q_0_x_Gauss}.

\textbf{EP update of $q_{x, 0}(\pmb x)$:} The first step of any EP update based on Gaussian approximations consists of computing the mean and covariance matrix of the tilted distribution. Since $f_x(\pmb x|\pmb \theta)$ forms a product of independent GMMs and since $q_{x, 1}(\pmb x)$ is a multivariate Gaussian density with a block-diagonal covariance matrix matching the covariance structure of $f_x(\pmb x|\pmb \theta)$, the tilted distribution $P_0(\pmb x) =f_x(\pmb x|\pmb \theta)q_{x,1}(\pmb x)$ is also a product of $J$ independent GMMs, whose parameters can be computed analytically and independently. For the $j$-th image patch, the mean ${\mathbb E}_{P_0}[ \pmb x_j]$ and covariance matrix ${\rm Cov}_{P_0}(\pmb x_j)$ are given by
\begin{equation}
\mathbb{E}_{P_0}[\pmb x_j] = \sum\limits_{k=1}^K \hat\omega_{j, k} \hat{\pmb \mu}_{j, k}, \quad{\rm Cov}_{P_0}(\pmb x_j) = \sum\limits_{k=1}^K \hat\omega_{j, k} (\hat{\pmb \mu}_{j, k} \hat{\pmb \mu}_{j, k}^T + \widehat{\bf C}_{j, k}) - \mathbb{E}_{P_0}[\pmb x_j]\mathbb{E}^T_{P_0}[\pmb x_j],
\label{Eq: titled_q0}
\end{equation}
where $\hat{\omega}_{j, k}$, $\hat{\pmb \mu}_{j, k}$ and  $\widehat{\bf C}_{j,k}$ are obtained by
\begin{equation}
\begin{cases}
\hat{\omega}_{j,k} = \frac{\omega_k \mathcal{N}\left(\left[\pmb m_{x, 1}\right]_j; \tilde{\pmb \mu}_k,\left[{\bf \Sigma}_{x,1}\right]_j + \widetilde{\bf C}_k\right)}{\sum\limits_{k=1}^K  \omega_k \mathcal{N}\left(\left[\pmb m_{x, 1}\right]_j; \tilde{\pmb \mu}_k,\left[{\bf \Sigma}_{x,1}\right]_j+\widetilde{\bf C}_k\right)}, \\
\widehat{\bf C}_{j,k} = \left(\left[{\bf \Sigma}_{x,1}\right]_j^{-1} +\widetilde{\bf C}_k^{-1}\right)^{-1},\\
\hat{\pmb \mu}_{j, k} = \widehat{\bf C}_k \left(\left[{\bf \Sigma}_{x,1}\right]_j^{-1} \left[\pmb m_{x, 1}\right]_j +  \widetilde{\bf C}_k^{-1} \tilde{\pmb \mu}_k\right), 
\end{cases}
\label{Eq: patch-based tilted GMM}
\end{equation}
where $\left[{\bf \Sigma}_{x,1}\right]_j$ denotes the block on the diagonal of ${\bf \Sigma}_{x,1}$ associated with the marginal covariance matrix of the $j$-th patch and $\left[\pmb m_{x, 1}\right]_j$ corresponds to the mean of that patch (extracted from $\pmb m_{x, 1}$). Once the mean and covariance matrix of $P_0(\pmb x)$ are computed, the next step consists of the actual update of $({\pmb m}_{x, 0}, {\mathbf \Sigma}_{x, 0})$. Again, this can be done effectively by leveraging the block structure of the covariance matrices, as the blocks can be processed independently. Let ${\mathbf \Omega}_0 = {\mathbf \Sigma}_{x,0}^{-1}$, ${\mathbf \Omega}_1 = {\mathbf \Sigma}_{x,1}^{-1}$ be the precision matrices of $q_{x,0}(\pmb x)$ and $q_{x,1}(\pmb x)$, respectively. These matrices can be easily obtained by $J$ inversions of small matrices of size $r\times r$, ${\mathbf \Omega}_0 = {\textrm diag}([{\mathbf \Sigma}_{x,0}]_1^{-1},\dots, [{\mathbf \Sigma}_{x,0}]_J^{-1})$, ${\mathbf \Omega}_1 = {\textrm diag}([{\mathbf \Sigma}_{x,1}]_1^{-1},\dots, [{\mathbf \Sigma}_{x,1}]_J^{-1})$. The KL divergence in \eqref{Eq: q_0_x_Gauss} can be expressed as
\begin{equation}
    KL(P_0(\pmb x)||Q(\pmb x)) = \displaystyle\int P_0(\pmb x) \left[ -\frac{1}{2}\log({\rm det}({\mathbf \Omega}_*)) +\frac{1}{2}(\pmb x-\pmb m_*)^T{\mathbf \Omega}_*(\pmb x- \pmb m_*) \right]{\rm d}{\pmb x} +\kappa,
\label{Eq: KL_P0_Q}
\end{equation}
where ${\mathbf \Omega}_* = {\mathbf\Omega}_0 +{\mathbf \Omega}_1$ and $\kappa=\int P_0(\pmb x)\log P_0(\pmb x){\rm d}{\pmb x}+\frac{N}{2}\log(2\pi)$ is a constant independent of $Q(\pmb x)$. Minimizing \eqref{Eq: KL_P0_Q} is 
equivalent to minimizing the loss function $F_0(\pmb m_*,{\mathbf \Omega}_*)$ defined as
\begin{equation}
F_0(\pmb m_*,{\mathbf \Omega}_*) = -\log ({\rm det}{(\bf \Omega}_*)) +\langle{\bf \Omega}_*, {\rm Cov}_{P_0}(\pmb x)\rangle +\left({\mathbb E}_{P_0}[\pmb x] - \pmb m_*\right)^T{\bf \Omega}_*\left({\mathbb E}_{P_0}[\pmb x] - \pmb m_*\right),
\label{Eq: loss_func_mStar_OmegaStar}
\end{equation}
with $\langle {\bf \Omega}_*, {\rm Cov}_{P_0}(\pmb x) \rangle = {\rm Trace}({\bf \Omega}_* {\rm Cov}_{P_0}(\pmb x))$. $F_0(\pmb m_*, {\bf \Omega}_*)$ is convex with respect to (w.r.t.) $\pmb m_*$ and minimized when $\pmb m_* = {\mathbb E}_{P_0}[\pmb x]$ (no constraints are applied to $\pmb m_*$). In that case, the loss function (as a function of ${\mathbf \Omega}_*$) reduces to 
\begin{equation}
    F_1({\mathbf \Omega}_*) =-(\log (\det {\mathbf \Omega}_*)) +\langle{\bf \Omega}_*, {\rm Cov}_{P_0}(\pmb x)\rangle.
\end{equation}
By taking advantage of the block-diagonal structure of the covariance matrix ${\rm Cov}_{P_0}(\pmb x)$ of the tilted distribution and of ${\mathbf \Sigma}_{x,1}$, $F_1({\mathbf\Omega}_*)$ can be re-written as
\begin{equation}
F_1({\mathbf \Omega}_0) \propto \sum\limits_{j=1}^J -\log(\det({\mathbf \Omega}_{0,j}+{\mathbf{\Omega}_{1,j}})) + \langle({\mathbf \Omega}_{0,j}+{\mathbf \Omega}_{1,j}), {\rm Cov}_{P_0}(\pmb x_j)\rangle,
\label{Eq: loss_func_Omega0}
\end{equation}
which can be minimized patch-wise w.r.t. $\{{\mathbf \Omega}_{0,j}\}_{j=1,\dots,J}$ via gradient-based methods. Here, we use an iterative method, where at the $(t)$-th iteration, we have
\begin{equation}
{\mathbf \Omega}_{0,j}^{(t)} = {\mathbf \Omega}_{0,j}^{(t-1)}-\lambda^{(t)}[{\rm Cov}_{P_0}(\pmb x_j) -({\mathbf \Omega}_{0,j}^{(t-1)} + {\mathbf \Omega}_{1,j})^{-1}].
\label{Eq: update_diag_BD}
\end{equation}
At each iteration, the initial step-size is chosen by Barzilai-Borwein method \cite{barzilai1988two} and it is then adjusted by backtracking line search to make sure the KL divergence decreases between successive iterations and that ${\mathbf \Omega}_{0,j}^{(t)} \in \mathcal{S}_{++}^r$, where $\mathcal{S}_{++}^r$ denotes the space of $r\times r$ symmetric, positive definite matrices. Once ${\mathbf \Omega}_{0,j}$ is obtained, the $j$-th diagonal block of ${\mathbf \Sigma}_{x,0}$ and $j$-th subset of $\pmb m_{x,0}$ are updated using
\begin{equation}
[{\mathbf \Sigma}_{0,x}]_j =  {\mathbf \Omega}_{0,j}^{-1}, \quad\quad [{\pmb m}_{x,0}]_j = [{\mathbf \Sigma}_{0,x}]_j \left(\left({\mathbf \Omega}_{0,j}+{\mathbf \Omega}_{1,j}\right){\mathbb E}_{P_0}[\pmb x_j]-{\mathbf \Omega}_{1,j}[\pmb m_{x,1}]_j\right).
\label{Eq: update_subsets_of_m0}
\end{equation}

\textbf{EP update of $q_{x,1}(\pmb x)$:} In \eqref{Eq: q_1_x_Gauss}, the tilted distribution $P_1(\pmb x)\propto\prod\limits _{n=1}^N {\mathcal N}(y_n;\pmb h_n\pmb x,\sigma^2)q_{x, 0}(\pmb x)$ is the product of two Gaussian densities and is thus also a multivariate Gaussian density. Its mean $\mathbb{E}_{P_1}[\pmb x]$ and covariance matrix ${\rm{Cov}}_{P_1}[\pmb x]$ are given by
\begin{equation}
\mathbb{E}_{P_1}[\pmb x] = {\rm{Cov}}_{P_1}(\pmb x)\left(\sigma^{-2} {\bf H}^T \pmb y+{\bf \Sigma}_{x, 0}^{-1}\pmb m_{x,0}\right), \quad \quad {\rm Cov}_{P_1}(\pmb x) = (\sigma^{-2}{\bf H}^T{\bf H}+{\bf \Sigma}_{x, 0}^{-1})^{-1}.
\label{Eq: titled_q1_Gaussian}
\end{equation}
The main computational bottleneck of this step is the inversion an $N\times N$ matrix required to compute ${\rm Cov}_{P_1}(\pmb x)$, and in turn $\mathbb{E}_{P_1}[\pmb x]$. If ${\bf H}^T{\bf H}$ is diagonal, the inversion can be computed patch-wise using the structure of ${\bf \Sigma}_{x, 0}^{-1}$. If ${\bf H}^T{\bf H}$ is non-diagonal, the direct inversion of $\sigma^{-2}{\bf H}^T{\bf H}+{\bf \Sigma}_{x, 0}^{-1}$ rapidly becomes too costly. To tackle this problem, the mean of the tilted distribution ${\mathbb E}_{P_1}[\pmb x]$ is computed by solving the following problem
\begin{eqnarray}
{\mathbb E}_{P_1}[\pmb x]= \underset{\pmb z}{\textrm{argmin}}~~\norm{{\rm Cov}_{P_1}^{-1}(\pmb x) {\pmb z} - \left(\sigma^{-2}{\bf H}^T\pmb y+ {\bf \Sigma}^{-1}_{x, 0} \pmb m_{x, 0} \right)}_2^2,
\label{Eq: E_P1_x}
\end{eqnarray}
since computing ${\rm Cov}_{P_1}^{-1}(\pmb x)$ is not as costly. This is achieved via conjugate gradient descent, which is a classical choice to solve such problems \cite{hestenes1952methods}. We have shown in \eqref{Eq: loss_func_Omega0} that only the blocks on the main diagonal of ${\rm Cov}_{P_i}(\pmb x)$ are required to minimize the KL divergence when the approximating distribution $Q(\pmb x)$ has a block-diagonal covariance matrix. Instead of inverting $\sigma^{-2}{\bf H}^T{\bf H}+{\bf \Sigma}_{x, 0}^{-1}$ and selecting those blocks, we approximate those marginal blocks using Rao–Blackwellized Monte Carlo (RBMC) \cite{siden2018efficient}, which provides in practice sufficiently accurate approximations.  We refer interested readers to \cite{siden2018efficient} for a detailed analysis about the time and memory complexity of RBMC. Once the $J$ diagonal blocks of ${\rm Cov}_{P_1}(\pmb x)$, i.e., $\{{\rm Cov}_{P_1}(\pmb x_j)\}_{j=1,\ldots,J}$ are estimated, $(\pmb m_{x,1},{\mathbf \Sigma}_{x,1})$ can be updated in a similar fashion to $(\pmb m_{x,0},{\mathbf \Sigma}_{x,0})$. For the precision matrix of the $j$-th block, i.e., ${\mathbf \Omega}_{1,j} = [{\mathbf \Sigma}_{x,1}]_j^{-1}$, the $(t)$-th gradient descent iteration reduces to
\begin{equation}
{\mathbf \Omega}_{1,j}^{(t)} = {\mathbf \Omega}_{1,j}^{(t-1)}-\lambda^{(t)}[{\rm Cov}_{P_1}(\pmb x_j) -({\mathbf \Omega}_{1,j}^{(t-1)} + {\mathbf \Omega}_{0,j})^{-1}].
\label{Eq: update_block_cov_of_q1}
\end{equation}
Once ${\mathbf \Omega}_{1,j}$ is obtained, the $j$-th diagonal block of ${\mathbf \Sigma}_{x, 1}$ and $j$-th subset of $\pmb m_{x, 1}$ are updated by
\begin{equation}
[{\mathbf \Sigma}_{x, 1}]_j = {\mathbf \Omega}^{-1}_{1,j}, \quad \quad [{\pmb m}_{x, 1}]_j = [{\mathbf \Sigma}_{x, 1}]_j\left(({\mathbf \Omega}_{1,j}+{\mathbf \Omega}_{0,j}){\mathbb E}_{P_1}[\pmb x_j]-{\mathbf \Omega}_{0,j}[\pmb m_{x, 0}]_j\right).
\label{Eq: update_subset_mean_of_q1}
\end{equation}
The same as for the update of $q_{x,0}(\pmb x)$, the $J$ blocks of ${\mathbf \Sigma}_{x, 1}$ and $J$ subsets of $\pmb m_{x, 1}$ can be updated independently in parallel.

While ${\mathbf \Sigma}_{x,0}$ and ${\mathbf \Sigma}_{x,1}$ are constrained to be block-diagonal in \eqref{Eq: update_subsets_of_m0} and \eqref{Eq: update_subset_mean_of_q1}, we can easily impose more restrictive constraints and force these matrices to be diagonal. For instance, in \eqref{Eq: update_subset_mean_of_q1}, if ${\mathbf \Omega}_{0,j}$ is non-diagonal and we want to ensure that ${\mathbf \Omega}_{1,j}$ is diagonal, it is sufficient to keep only the diagonal of the gradient term in \eqref{Eq: update_block_cov_of_q1}, provided that the initial ${\mathbf \Omega}_{1,j}^{(1)}$ is diagonal. In practice, as will be shown in Section \ref{Sec: Experimental Results}, we have observed that constraining ${\mathbf \Sigma}_{x,0}$ and ${\mathbf \Sigma}_{x,1}$ to be diagonal leads to satisfactory results for denoising and inpainting problems where ${\bf H}^T{\bf H}$ is diagonal, and the resulting algorithm is faster than if ${\mathbf \Sigma}_{x,0}$ and ${\mathbf \Sigma}_{x,1}$ are constrained to be only block-diagonal (See Appendix \ref{subsec: appen_diagonal covariance} for details). Moreover, if ${\bf H}^T{\bf H}$ is diagonal and positive-definite (e.g., $\textbf{H}$ is the identity matrix), the gradient descent method in \eqref{Eq: update_block_cov_of_q1} is not needed and we can directly set ${\mathbf \Sigma}_{x,1}=\sigma^2({\bf H}^T{\bf H})^{-1}$. Conversely, we have observed that forcing ${\mathbf \Sigma}_{x,0}$ and ${\mathbf \Sigma}_{x,1}$ to be diagonal for deconvolution problems do not lead to acceptable results in general as in this case, the resulting EP message passing structure in \eqref{Eq: KL minimization for approx Gaussian likeli} does not capture nor pass on important correlations originally included in the exact posterior. The pseudo-code of the proposed EP algorithm for Gaussian observation model is presented in Algorithm~\ref{alg: algorithm_Gaussian_likeli}.

\begin{algorithm}[!ht]
\caption{Proposed patch-based EP algorithm - Gaussian observation model}
\label{alg: algorithm_Gaussian_likeli}
\begin{algorithmic}[1]
\Require $\pmb y$, $\bf H$, $\sigma^2$, \{$\omega_k$, $\tilde{\pmb \mu}_k$, $\widetilde{\bf C}_k$, $K$\} 
\Ensure ${\bf \Sigma}_{x, 0}$, $\pmb m_{x, 0}$, ${\bf \Sigma}_{x, 1}$, $\pmb m_{x, 1}$\\
initialization: $\{\pmb m_{x,1},\pmb m_{x,0}\} =\pmb y$, $\{{\mathbf \Sigma}_{x,1},{\mathbf \Sigma}_{x,0}\} = \sigma^2 {\mathbf I}$\;
\While {stopping criterion is not satisfied}
\State \textbullet~\textbf{EP update of $q_{x, 0}(\pmb x)$} 
\Indent
\State compute titled mean and block covariance via \eqref{Eq: titled_q0}
\For {$j=1,\dots,J$ in parallel}
\State update the precision ${\mathbf \Omega}_{0,j}$ of diagonal blocks
\State update the covariance of diagonal block $[{\mathbf \Sigma}_{x,0}]_j$ and subset of mean $[\pmb m_{x, 0}]_j$
\EndFor 
\EndIndent
\State \textbullet~\textbf{EP update of $q_{x, 1}(\pmb x)$} 
\Indent
\State  compute titled mean and block covariance by RBMC in \eqref{Eq: titled_q1_Gaussian} 
\For{$j=1,\dots,J$ in parallel}
\State update the precision ${\mathbf \Omega}_{1,j}$ of diagonal blocks
\State update the covariance of diagonal block $[{\mathbf \Sigma}_{x, 1}]_j$ and subset of mean $[\pmb m_{x, 1}]_j$
\EndFor
\EndIndent
\EndWhile
\end{algorithmic}
\end{algorithm}

\section{Patch-based EP algorithm with non-Gaussian noise}
\label{Sec: Patch-based EP with non-Gaussian noise}

\subsection{Poisson observation model}
For restoration of images corrupted by Poisson noise, the observations $y_n$ for $n=1,\dots,N$ are classically assumed to be mutually independent, conditioned on $\pmb x$, and the likelihood can be expressed as 
\begin{eqnarray}
f_{y|x}(\pmb y|{\mathbf H} \pmb x) = \prod\limits_{n=1}^N {\mathcal P}_{y_n}(\pmb h_n\pmb x) = \prod\limits_{n=1}^N \frac{e^{-\pmb h_n\pmb x}(\pmb h_n\pmb x)^{y_n}}{y_n!},
\end{eqnarray}
where the entries of $\{\pmb h_n\}_{n=1,\dots,N}$ are non-negative. This likelihood is particularly challenging as  (\romannumeral 1) it usually requires $\pmb x$ to be positive, and (\romannumeral 2) it induces noise levels that change across the $N$ observations depending on ${\mathbf H}$ and $\pmb x$. This makes the Poisson model an interesting case study for the quantification of posterior uncertainty. 

Imposing positivity constraints on $\pmb x$ via its prior model requires modifying the GMM in \eqref{Eq: proposed GMM patch-based prior}. However, introducing mixtures of multivariate truncated Gaussian distributions is not a suitable option as the normalizing constants involved in the computation of the mixture weights would be intractable. To alleviate such issues, we still consider the GMM prior model \eqref{Eq: proposed GMM patch-based prior} but extend the Poisson distribution using the rectified linear transform \cite{ko2016expectation}. More precisely, $\forall u \in \mathbb{R}$, we use the data likelihood
\begin{eqnarray}
\bar{\mathcal{P}}_y(u) =
\frac{u^{y}e^{-u}}{y!}\mathbb{I}(u>0) + \delta(y)\mathbb{I}(u \leq 0),
\label{Eq: Rectified_Poisson_likeli}
\end{eqnarray}
where ${\mathbb I}(.)$ is an indicator function and $\delta(\cdot)$ is the Dirac delta function. For $u>0$, $\bar{\mathcal{P}}_y(u)$ reduces to the standard p.m.f. of the Poisson distribution (with positive mean), while $y$ is forced to be $0$ when $u\leq0$. Combining \eqref{Eq: Rectified_Poisson_likeli} with the GMM patch-based prior defined in \eqref{Eq: proposed GMM patch-based prior}, the exact posterior is given by
\begin{equation}
f(\pmb x|\pmb y,\pmb \theta) \propto \left[\prod\limits _{n=1}^N {\bar{\mathcal P}}_{y_n}(\pmb h_n\pmb x)\right]f_x(\pmb x|\pmb \theta).
\label{Eq: exact_posterior_Poisson}
\end{equation}

With this exact model, if we choose the same EP factorization as in Section \ref{Sec: patch-basedEP_Gaussian_observation}, ${\mathbb  E}_{P_1}[\pmb x]$ and the diagonal blocks of ${\rm Cov}_{P_1}(\pmb x)$ cannot be computed easily due to combination of the Poisson likelihood and the linear operator $\mathbf H$. Instead, we introduce a data augmentation scheme which fits conveniently within the EP framework and allows for simpler updates. 

\subsection{Data augmentation and EP with Poisson noise} 
To decouple the non-Gaussian (Poisson) noise from the linear operator $\mathbf H$, we extend the Bayesian model in \eqref{Eq: exact_posterior_Poisson} to 
\begin{equation}
f(\pmb u,\pmb x|\pmb y,\pmb \theta) \propto f(\pmb y|\pmb u)f(\pmb u|\pmb x)f_x(\pmb x|\pmb \theta),
\label{Eq: extended posterior using u}   
\end{equation}
where $f(\pmb y|\pmb u) = \prod\nolimits_{n=1}^N \bar{{\mathcal P}}_{y_n}(u_n)$ and $f(\pmb u|\pmb x) = \delta(\pmb u - {\mathbf H}\pmb x)=\prod_{n=1}^N\delta(u_n-\pmb h_n \pmb x)$ \cite{kim2018expectation}. Note that the marginal distribution obtained by integrating $f(\pmb u,\pmb x|\pmb y,\pmb \theta)$ w.r.t. $\pmb u$ is the original posterior in \eqref{Eq: exact_posterior_Poisson}. Instead of approximating \eqref{Eq: exact_posterior_Poisson} using $Q(\pmb x)$, we will approximate the extended posterior distribution \eqref{Eq: extended posterior using u} using a distribution $Q(\pmb x, \pmb u)$, whose marginal (after integration over $\pmb u$) will approximate $f(\pmb x|\pmb y,\pmb \theta)$ in \eqref{Eq: exact_posterior_Poisson}.

\begin{figure}[!ht]
\centering
\includegraphics[width=0.8\textwidth]{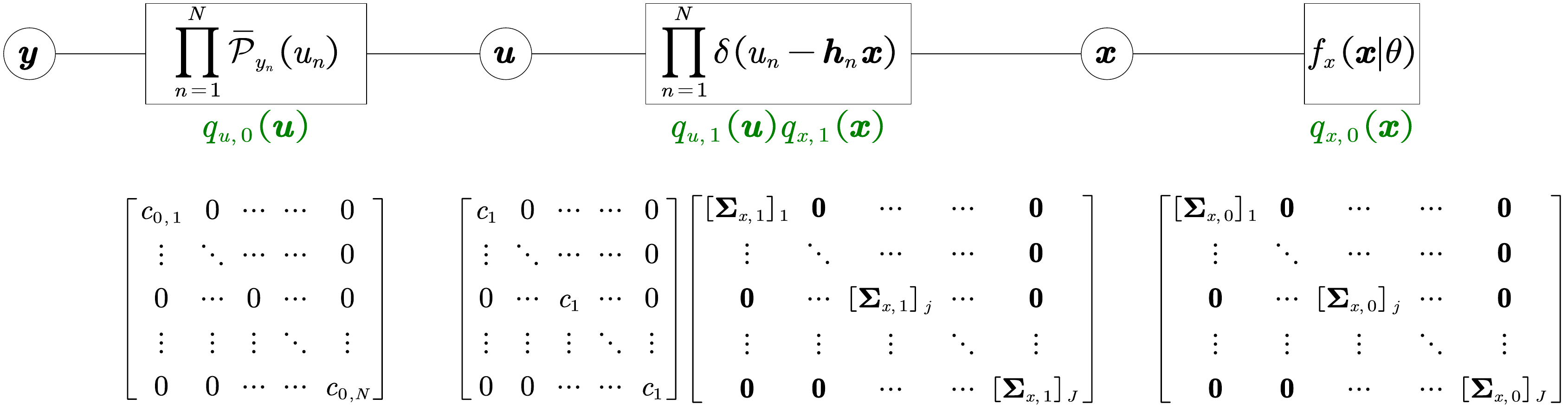}
\caption{Factor graph used to perform EP approximation for Poisson noise model using data augmentation. The rectangular boxes (resp. circles) represent the factors (resp. variables) involved in the factorized exact posterior distribution. The corresponding EP approximating factors are shown in green.}
\label{fig: factor_graph_2}
\end{figure}

While the model in \eqref{Eq: exact_posterior_Poisson} naturally factorizes into two factors, the extended model in \eqref{Eq: extended posterior using u} naturally factorizes using three factors due to the hierarchical structure $\pmb y - \pmb u - \pmb x$. The first factor $f(\pmb y|\pmb u)$ only depends on $\pmb u$ (and $\pmb y$ which is observed) and will be approximated by a Gaussian density denoted $q_{u, 0}(\pmb u)$. The third factor $f_x(\pmb x|\pmb \theta)$ only depends on $\pmb x$ (and $\pmb \theta$ which is fixed for now) and will be approximated by $q_{x, 0}(\pmb x)$, as in Section \ref{subsec: EP update for block-diagonal covariance with positive definite constriant}. The second factor $f(\pmb u|\pmb x)$ depends on $\pmb u$ and $\pmb x$ and is approximated by a multivariate Gaussian density $q_1(\pmb u,\pmb x)$ too. Although  $\pmb u$ and $\pmb x$ are highly and explicitly correlated in the exact model \eqref{Eq: extended posterior using u}, we use a mean-field approximation and specify $q_1(\pmb u,\pmb x)=q_{u,1}(\pmb u)q_{x,1}(\pmb x)$, where $q_{u,1}(\cdot)$ and $q_{x,1}(\cdot)$ are two Gaussian densities. Such approximations are common in VB methods and here, it still allows the approximated posterior means of $\pmb u$ and $\pmb x$ to be implicitly correlated. It also enables simpler EP updates, as will be shown next. The resulting factor graph and EP factorization are depicted in Figure \ref{fig: factor_graph_2}, where the approximating factors are characterized by
\begin{equation}
  q_{u, i}(\pmb u)\propto {\mathcal N}(\pmb u;
\pmb m_{u, i},{\bf \Sigma}_{u, i})  \quad \textrm{and}\quad  q_{x, i}(\pmb x)\propto {\mathcal N}(\pmb x;
\pmb m_{x, i},{\bf \Sigma}_{x, i}), \quad  \forall i \in \{0,1\} .
\end{equation}

Note that using the mean-field approximation, the approximating distribution $Q(\pmb u, \pmb x) \propto q_{u,0}(\pmb u)q_{u,1}(\pmb u)q_{x,1}(\pmb x)q_{x,0}(\pmb x)$ can be written as $Q(\pmb u, \pmb x)=Q(\pmb u)Q(\pmb x)$, which gives us directly access to the moments of the marginal distribution $\int Q(\pmb u, \pmb x)\textrm{d}\pmb u$ that will be used as surrogate for the intractable moments of $f(\pmb x|\pmb y,\pmb \theta)$ in \eqref{Eq: exact_posterior_Poisson}.
Using these three factors, each EP iteration consists of sequentially solving the three following KL divergence minimization problems
\begin{subnumcases}{\hspace{1cm}\label{Eq: KL minimization for approx Poisson likeli_extended}}
\hspace{0.4cm}\mathop{\min}\limits_{q_{u, 0}(\pmb u)} \hspace{0.6cm}KL\left(\prod\nolimits_{n=1}^N \bar{{\mathcal P}}_{y_n}(u_n)q_{u, 1}(\pmb u)||q_{u, 0}(\pmb u)q_{u, 1}(\pmb u)\right),\label{Eq: q_1_u}\\
\mathop{\min}\limits_{q_{u, 1}(\pmb u),q_{x, 1}(\pmb x)} KL\left(\prod_{n=1}^N\delta(u_n-\pmb h_n \pmb x)q_{u, 0}(\pmb u)q_{x, 0}(\pmb x)||q_{u, 1}(\pmb u)q_{x, 1}(\pmb x)q_{u, 0}(\pmb u)q_{x, 0}(\pmb x)\right),\label{Eq: q_0_u_q_1_x}\\
\hspace{0.5cm}\mathop{\min}\limits_{q_{x, 0}(\pmb x)} \hspace{0.53cm}KL\left(f_{x}(\pmb x|\pmb \theta)q_{x, 1}(\pmb x)||q_{x, 0}(\pmb x)q_{x, 1}(\pmb x)\right), \label{Eq: q_0_x}
\end{subnumcases}
subject to structural constraints on different covariance matrices of the approximating factors.

\subsection{EP updates with data augmentation}
As in the Gaussian noise case, structural constraints can be enforced on the covariance matrices in the case of non-Gaussian noise with extended models. For vector $\pmb x$, it makes sense to consider the same constraints as in the Gaussian case since $f_{x}(\pmb x|\pmb \theta)$ is still a patch-based prior and $f(\pmb u|\pmb x)$ conveys correlation between pixels via ${\bf H}$. Thus, we constrain ${\bf \Sigma}_{x, 0}$ and ${\bf \Sigma}_{x, 1}$ to be block-diagonal matrices, with the same block structure as the GMM prior. The actual likelihood term $f(\pmb y|\pmb u)$ does not convey correlation between the elements of $\pmb u$ since $f(\pmb y|\pmb u) = \prod\nolimits_{n=1}^N \bar{{\mathcal P}}_{y_n}(u_n)$. Thus, we naturally constrain ${\bf \Sigma}_{u, 0}$ to be diagonal and its diagonal is denoted by $[c_{0,1},\dots,c_{0,N}]$. Another option would be to enforce ${\bf \Sigma}_{u, 0}$ to be isotropic but this would not capture accurately the pixel-wise uncertainty induced by the Poisson likelihood.  This can be a valid option for other types of noise such as zero-mean, Laplace-distributed noise, identically distributed across all the observations. The remaining covariance matrix to be considered is ${\bf \Sigma}_{u, 1}$. In practice, allowing too much flexibility can make EP updates unstable and although we could force this matrix to be diagonal, forcing it to be isotropic leads to much more stable results. Thus, we define ${\bf \Sigma}_{u, 1}=c_{1}{\mathbf I}$, where ${\mathbf I}\in {\mathbb R}^{N\times N}$ denotes an identity matrix. An extended discussion about constraining covariance matrices in EP is included in Section \ref{Sec: Discussion and Conclusions}. Once the constraints on different covariance matrices are set, we can discuss how the problems in \eqref{Eq: KL minimization for approx Poisson likeli_extended} can be solved. Since \eqref{Eq: q_0_x} is the same problem as \eqref{Eq: q_0_x_Gauss}, we focus on \eqref{Eq: q_1_u} and \eqref{Eq: q_0_u_q_1_x}.

\textbf{EP update of $q_{u, 0}(\pmb u)$:} In \eqref{Eq: q_1_u}, we need to compute the first and second-order moments of the tilted distribution $P_0(\pmb u) = \prod\nolimits_{n=1}^N \bar{{\mathcal P}}_{y_n}(u_n)q_{u, 1}(\pmb u)$. Fortunately, this distribution factorizes over the $N$ elements of $\pmb u$ since $q_{1,u}(\pmb u)$ has an isotropic (and thus diagonal) covariance matrix. Consequently, we only need to compute the marginal means and variances associated with $P_0(\pmb u)$, denoted by ${\mathbb E}_{P_0}[u_n]$ and ${\rm Var}_{P_0}[u_n]$. These moments can be computed using only 1D integrals by introducing the marginal tilted distributions $P_0(u_n)= \bar{\mathcal P}_{y_n}(u_n)\mathcal N(u_n; \mu_{1,n},c_1)$, where we use $\mu_{1,n}:=m_{u,1,n}, \forall n \in N$ to simplify the notation. 

If $y_n = 0$, then $\bar{\mathcal{P}}_{y_n=0}(u_n)$ reduces to $e^{-u_n}\mathbb{I}(u_n>0) + \mathbb{I}(u_n\leq 0)$ and the tilted distribution $ P_0(u_n)$ is a mixture of two truncated Gaussian distributions, whose mean and variance can be computed in closed-form \cite[Chap.~2]{ko2017applications}. On the other hand, if $y_n \neq 0$, computing the mean and variance of $P_0(u_n)$ is less trivial but still possible by using only the following 1D integrals
\begin{equation}
\begin{cases}
   Z_n = \int P_0(u_n) {\rm d}{u_n},\\
   \mathbb{E}_{P_0}[u_n] = \frac{1}{Z_n}\int u_n P_0(u_n) {\rm d}{u_n},\\
   \mathbb{E}_{P_0}[u_n^2] = \frac{1}{Z_n}\int u_n^2 P_0(u_n) {\rm d}{u_n}.
\end{cases}
\label{Eq: moments_of_tilted_for_q1_u}
\end{equation}
Two main approaches can be adopted to compute these integrals. Ko et al. reported recursive formulas to compute the first two moments of $P_0(u_n)$ at the cost of $\mathcal{O}(y_n+2)$ in \cite{ko2016expectation,ko2017applications}. However, these recursive formulas were found to be numerically unstable with moderately high values of $y_n$. In this work, we use the quadrature-based implementation proposed in \cite[Appendix~B]{wand2011mean}, which presents a higher computational complexity but is more stable. Such a quadrature method can also be used for other noise models, as long as the observations in $\pmb y$ are mutually independent given $\pmb u$. In such cases, it is sufficient to find approximate solutions to compute \eqref{Eq: moments_of_tilted_for_q1_u}. For instance, in first-photon or low-photon imaging applications, the observations can be assumed to follow binomial or geometric distributions\cite{altmann2017unsupervised}, and the proposed EP framework could be applied to such problems without major modifications.

Once the marginal means $\{{\mathbb E}_{P_0}[u_n]\}_{n=1,\dots,N}$ and variances $\{{\rm Var}_{P_0}[u_n]\}_{n=1,\dots,N}$ are computed, the mean and marginal variances of $q_{u,0}(\pmb u)$ can be updated element-wise via
\begin{equation}
\frac{1}{c_{0,n}} = \frac{1}{{\rm Var}_{P_0}[u_n]} - \frac{1}{c_1},\quad\quad \mu_{0,n} = c_{0,n}\left(\mathbb{E}_{P_0}[u_n]\left(\frac{1}{c_{0,n}}+\frac{1}{c_1}\right) - \frac{\mu_{1,n}}{c_1}\right),
\label{Eq: update_var_and_mean_q1_u} 
\end{equation} 
using the fact that $Q(\pmb u)$ has a diagonal covariance matrix (See Appendix \ref{subsec: appen_diagonal covariance}). Note that \eqref{Eq: update_var_and_mean_q1_u} does not ensure that $c_{0,n}$ is positive. In practice, if the value obtained is negative, it is replaced by assigning a large value (e.g., $10^8$) before computing $\mu_{0,n}$. 

\textbf{EP update of $q_{x,1}(\pmb x)$:}
In \eqref{Eq: q_0_u_q_1_x}, the joint tilted distribution is $P_1(\pmb u,\pmb x) = \prod_{n=1}^N\delta(u_n-\pmb h_n \pmb x)q_{u, 0}(\pmb u)q_{x, 0}(\pmb x)$. However, since the approximation $Q(\pmb u,\pmb x)$ assumes independence between $\pmb u$ and $\pmb x$, it is sufficient to compute the means and  marginal variances of $P_1(\pmb x) = \int P_1(\pmb u,\pmb x) {\rm d}{\pmb u}$ and $P_1(\pmb u) = \int P_1(\pmb u,\pmb x) {\rm d}{\pmb x}$ to update $q_{u, 1}(\pmb u)$ and $q_{x, 1}(\pmb x)$, respectively.

To update $q_{x,1}(\pmb x)$, we need to compute the moments of $P_1(\pmb x) \propto \int P_1(\pmb u,\pmb x) {\rm d}{\pmb u}$, which can be computed in closed-form using 
\begin{equation}
\begin{aligned}
P_1(\pmb x) & \propto  q_{x, 0}(\pmb x)\int \delta(\pmb u - {\bf H}\pmb x)q_{u, 0}(\pmb u) {\rm d}{\pmb u}\\
& \propto  q_{x, 0}(\pmb x)\mathcal{N}({\bf H}\pmb x;\pmb m_{u,0},{\bf \Sigma}_{u,0}).
\end{aligned}
\end{equation}
It follows that the mean ${\mathbb E}_{P_1}[\pmb x]$ and covariance ${\rm Cov}_{P_1}(\pmb x)$ are given by
\begin{equation}
{\rm Cov}_{P_1}(\pmb x) = ({\bf \Sigma}_{x, 0}^{-1} + {\bf H}^T{\bf \Sigma}_{u, 0}^{-1}{\bf H})^{-1}, \quad \quad \mathbb{E}_{P_1}[\pmb x] = {\rm Cov}_{P_1}(\pmb x)\left({\bf \Sigma}_{x, 0}^{-1}\pmb m_{x, 0} + {\bf H}^T {\bf \Sigma}_{u, 0}^{-1}\pmb m_{u,0}\right).
\label{Eq: marginal_P_x}
\end{equation} 
Since $Q(\pmb x)$ has a block-diagonal covariance matrix, the update of $\pmb m_{x, 1}$ and the blocks on the diagonal of ${\bf \Sigma}_{x, 1}^{-1}$ is performed the same as in Section \ref{subsec: EP update for block-diagonal covariance with positive definite constriant}, i.e., using the RBMC method.

\textbf{EP update of $q_{u, 1}(\pmb u)$:}  Since $Q(\pmb u)$ has a diagonal covariance matrix, we only need to compute the marginal moments of $P_1(u_n)\propto\int P_1(\pmb u) \textrm{d}\pmb u_{\backslash n}, \forall n \in N$, where $\pmb u_{\backslash n}$ denotes the elements of $\pmb u$ whose element $u_n$ has been removed. Using the formula provided in \cite{kim2018expectation} to compute the integral of the product of a Gaussian density by a Dirac delta function, we obtain
\begin{equation}
\begin{aligned}
P_1(u_n) & \propto \int \int P_1(\pmb u,\pmb x) \textrm{d}\pmb u_{\backslash n} \textrm{d} \pmb x\\
& \propto  \int \int \prod_{n=1}^N\delta(u_n-\pmb h_n \pmb x)q_{u, 0}(\pmb u)q_{x, 0}(\pmb x) \textrm{d}\pmb u_{\backslash n} \textrm{d} \pmb x\\
& \propto  q_{u,0}(u_n)\int \delta(u_n-\pmb h_n\pmb x)\int\delta(\pmb u_{\backslash n}-{\bf H}_{\backslash n}\pmb x)q_{u,0}(\pmb u_{\backslash n})\textrm{d}\pmb u_{\backslash n} q_{x, 0}(\pmb x)\textrm{d} \pmb x\\
& \propto  q_{u,0}(u_n)\int \delta(u_n-\pmb h_n\pmb x)q_{u,0}^{(\backslash n)}({\bf H}_{\backslash n}\pmb x) q_{x, 0}(\pmb x)\textrm{d} \pmb x\\
& \propto  q_{u, 0}(u_n){\mathcal N}(u_n;\pmb h_n \widetilde{\pmb m}_n,\pmb h_n \widetilde{\bf \Omega}_n^{-1}\pmb h_n^T).
\end{aligned}
\label{eq:marginal_qu1}
\end{equation}
${\mathbf H}_{\backslash n}$ is the matrix $\mathbf H$ whose $n$-th row has been removed, $q_{u,0}^{(\backslash n)}({\mathbf H}_{\backslash n}\pmb x) = {\mathcal N}({\mathbf H}_{\backslash n}\pmb x; \pmb m_{u,0}^{\backslash n},{\mathbf \Sigma}_{u,0}^{\backslash n})$, $\widetilde{\pmb m}_n = {\widetilde{\bf \Omega}}_n^{-1}\left[{\bf H}_{\backslash n}^T ({\bf \Sigma}_{u, 0}^{\backslash n})^{-1}\pmb m_{u, 0}^{\backslash n} \middlebreak + {\bf \Sigma}_{x, 0}^{-1}\pmb m_{x, 0}\right]$ and $\widetilde{\bf \Omega}_n = {\bf H}^T_{\backslash n}({\bf \Sigma}_{u,0}^{\backslash n})^{-1}{\bf H}_{\backslash n} +{\bf \Sigma}_{x, 0}^{-1}$. The last line of \eqref{eq:marginal_qu1}  shows that $P_1(u_n)$ is the product of two Gaussian densities. Thus its moments are given by
\begin{equation}
{\rm Var}_{P_1}(u_n) = \left(\dfrac{1}{c_{0,n}}+\dfrac{1}{\pmb h_n \widetilde{\bf \Omega}_n^{-1}\pmb h_n^T} \right)^{-1}, \quad 
{\mathbb E}_{P_1}[u_n]  = {\rm Var}_{P_1}(u_n) \left(\dfrac{\mu_{0,n}}{c_{0,n}}+\dfrac{\pmb h_n \widetilde{\pmb m}_n}{\pmb h_n \widetilde{\bf \Omega}_n^{-1}\pmb h_n^T}\right).
\label{eq:tilted_P1_u}
\end{equation}
The main computational cost here is the inversion of the large pixel-dependent matrix $\widetilde{\bf \Omega}_n^{-1}$ needed to compute $\pmb h_n \widetilde{\pmb m}_n$ and $\pmb h_n \widetilde{\bf \Omega}_n^{-1}\pmb h_n^T$. However, it can be noticed that ${\widetilde{\bf \Omega}}_n^{-1}$ is ${\rm Cov}_{P_1}(\pmb x)$ in \eqref{Eq: marginal_P_x} whose contribution of $u_n$ has been removed. Even for relatively small images, ${\widetilde{\bf \Omega}}_n^{-1} \approx {\rm Cov}_{P_1}(\pmb x)$ is a reasonable approximation. Instead of using this approximation which would still require the computation of ${\rm Cov}_{P_1}(\pmb x)$, we propose to use ${\widetilde{\bf \Omega}}_n^{-1} \approx {\bf \Sigma}_{*}$, which is a block-diagonal approximation of ${\rm Cov}_{P_1}(\pmb x)$ (computed during the update of $q_{x,1}(\pmb x)$ above). This allows a fast approximation of the marginal moments in \eqref{eq:tilted_P1_u}. Using these moments and $q_{u,0}(\pmb u)$, the updated $c_1$ can be obtained via gradient descent (or Newton-Raphson), as detailed in Appendix \ref{subsec: appen_covariance_iso}. After the update of $c_1$, $\pmb m_{u, 1}$ is obtained via
\begin{equation}
\pmb m_{u, 1} = c_1\left[\left(c_1^{-1}{\mathbf I}+{\mathbf \Sigma}_{u,0}^{-1}\right){\mathbb E}_{P_1}[\pmb u]-{\mathbf \Sigma}_{u, 0}^{-1}\pmb m_{u, 0}\right].
\label{Eq: update_of_q1_u}
\end{equation}

The pseudo-code of the EP algorithm for Poisson observation model is presented in Algorithm~\ref{alg: algorithm_Poisson_likeli}. The updates of $q_{u, 0}(\pmb u)$, $q_{x, 1}(\pmb x)$, $q_{u,1}(\pmb u)$, and $q_{x,0}(\pmb x)$ are repeated until some convergence conditions are met. Here, we use $||\pmb m_*^{(t)}-\pmb m_*^{(t-1)}||^2<1{\rm e}^{-8} N$ and $||{\rm diag}({\mathbf \Sigma}_*^{(t)})-{\rm diag}({\mathbf \Sigma}_*^{(t-1)})||^2<1{\rm e}^{-8}N$ as the stopping criteria ($(t)$ denotes the $t$-th EP iteration). In general, there are no convergence guarantees for EP algorithms. A damping strategy \cite{minka2005divergence, gelman2014expectation, hernandez2015expectation} is often used when oscillations between successive iterations are observed. Here, a damping with parameter $\epsilon=0.7$ is applied in Algorithms \ref{alg: algorithm_Gaussian_likeli} and \ref{alg: algorithm_Poisson_likeli}. After the last iteration, irrespective of the actual noise model, the EP approximation of the posterior $f(\pmb x|\pmb y,\pmb \theta)$ is given by $Q(\pmb x) \propto q_{x,0}(\pmb x)q_{x,1}(\pmb x)$, with mean $\pmb m_*$ and covariance matrix ${\bf \Sigma}_*$.

\begin{algorithm}[htbp]
\caption{Proposed patch-based EP algorithm - Poisson observation model}
\label{alg: algorithm_Poisson_likeli}
\begin{algorithmic}[1]
\Require $\pmb y$, $\bf H$, \{$\omega_k$, $\tilde{\pmb \mu}_k$, $\widetilde{\bf C}_k$, $K$\}
\Ensure ${\mathbf \Sigma}_{u,1}$, $\pmb m_{u,1}$, ${\mathbf \Sigma}_{u, 0}$, $\pmb m_{u, 0}$, ${\bf \Sigma}_{x,0}$, $\pmb m_{x, 0}$, ${\bf \Sigma}_{x, 1}$, $\pmb m_{x, 1}$\\
initialization: $\{\pmb m_{u,0},\pmb m_{u,1},\pmb m_{x,1},\pmb m_{x,0}\} =\pmb y+1$, $\{{\mathbf \Sigma}_{u,0},{\mathbf \Sigma}_{u,1},{\mathbf \Sigma}_{x,1},{\mathbf \Sigma}_{x,0}\} = {\rm diag}(\pmb y+1)$\; 
\While {stopping criterion is not satisfied}
\State \textbullet~\textbf{EP update of $q_{u, 0}(\pmb u)$} 
\Indent
\State compute titled covariance and mean of titled distribution in \eqref{Eq: q_1_u}
\State update the marginal variance and mean of $q_{u, 0}(\pmb u)$ via \eqref{Eq: update_var_and_mean_q1_u} 
\EndIndent
\State \textbullet~\textbf{EP update of $q_{x, 1}(\pmb x)$} 
\Indent
\State  compute titled mean and block covariance by RBMC in \eqref{Eq: titled_q1_Gaussian}
\For{$j=1,\dots,J$ in parallel}
\State update the precision ${\mathbf \Omega}_{1,j}$ of diagonal blocks
\State update the covariance of diagonal block $[{\mathbf \Sigma}_{x,1}]_j$ and subset of mean $[\pmb m_{x,1}]_j$
\EndFor
\EndIndent
\State \textbullet~\textbf{EP update of $q_{u, 1}(\pmb u)$} 
\Indent
\State compute the marginal moments of $\{P_1(u_n)\}_{n=1,\dots,N}$ in \eqref{eq:tilted_P1_u}
\State update the mean and variance of $q_{u, 1}(\pmb u)$ via \eqref{Eq: update_of_q1_u}
\EndIndent
\State \textbullet~\textbf{EP update of $q_{x, 0}(\pmb x)$} 
\Indent
\For{$j=1,\dots,J$ in parallel}
\State update the precision ${\mathbf \Omega}_{0,j}$ of diagonal blocks
\State update the covariance of diagonal block $[{\mathbf \Sigma}_{x,0}]_j$ and subset of mean $[\pmb m_{x,0}]_j$
\EndFor 
\EndIndent
\EndWhile
\end{algorithmic}
\end{algorithm}

\section{Final Estimation using Product-of-Experts}
\label{Sec: Final Estimation using Geometric Average}

\begin{figure}[t]
\centering
\centerline{\includegraphics[width=0.6\textwidth]{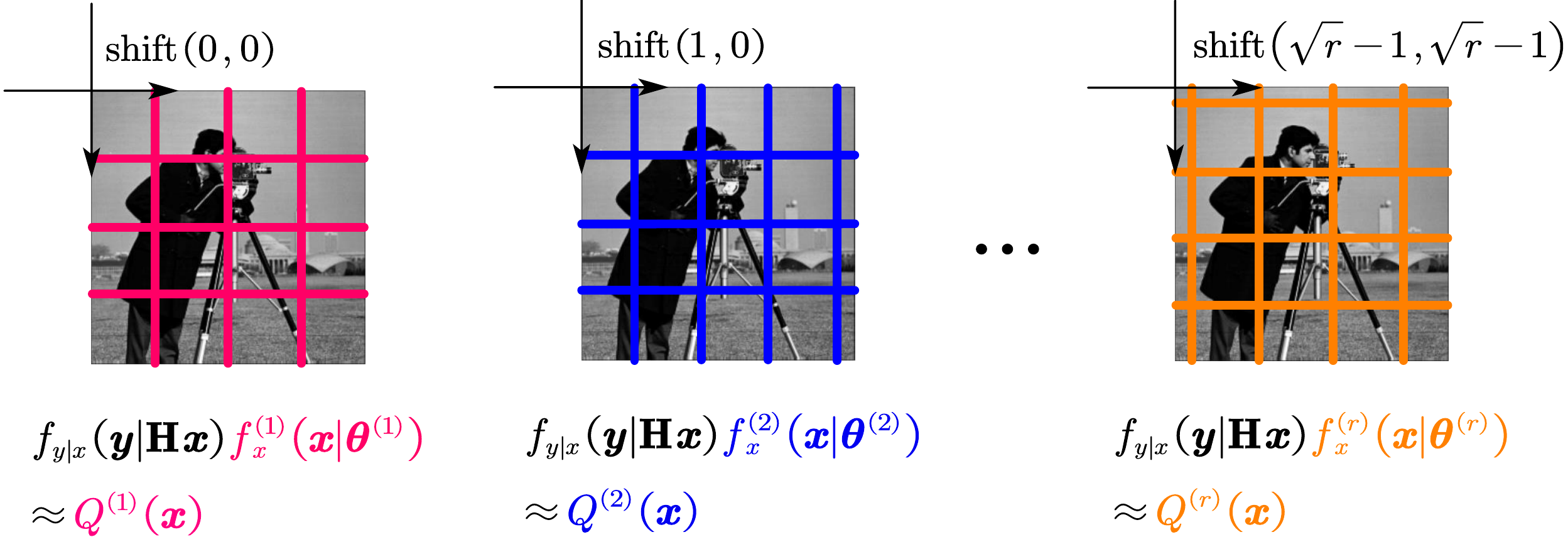}}
\caption{Illustration of a set of different GMM patch-based priors $f_x^{(i)}(\pmb x|\pmb \theta^{(i)})$ ($i=1,\dots,r$) modeling overlapping patches extracted from the same image. Approximation for exact posterior $f(\pmb x|\pmb y,\pmb \theta)$ is finally built as the Product-of-Experts $Q_f(\pmb x) \propto \left[Q^{(1)}(\pmb x)Q^{(2)}(\pmb x)\dots Q^{(r)}(\pmb x) \right]^{\frac{1}{r}}$.}
\label{fig: geometric_average}    
\end{figure}

Using the GMM prior in \eqref{Eq: proposed GMM patch-based prior}, the mean of $Q(\pmb x)$ presents blocky artefacts because the patches are non-overlapping and a priori mutually independent. To tackle this problem, instead of using a single prior and a unique partitioning of the pixels, we consider a set of prior models, with shifted partitions, as illustrated in Figure \ref{fig: geometric_average}.  Using vertical and horizontal steps of one pixel, we build $r$ distinct partitions, denoted ${\mathcal S}_1,\dots,{\mathcal S}_{r}$, leading to $r$ prior models $\{f^{(i)}_x(\pmb x|\boldsymbol{\theta}^{(i)})\}_{i=1,\ldots,r}$. For each of these models, the prior covariance matrix presents (after appropriate sorting of the pixels in $\pmb x$) a block-diagonal structure. Thus, the EP methods described in Sections \ref{Sec: patch-basedEP_Gaussian_observation} and \ref{Sec: Patch-based EP with non-Gaussian noise} can be used separately with any of these priors, which provide a set of $r$ approximations $Q^{(i)}(\pmb x) \approx f(\pmb x|\pmb y,\boldsymbol{\theta}^{(i)})$. These distributions can be combined in different ways. Model averaging could be achieved by assigning a prior probability (denoted $f({\mathcal S}_i)$ e.g., $f({\mathcal S}_i)=1/r$) to each partitioning of the pixels, and by marginalising over the $r$ partitions \cite[Chapter~7.4]{gelman2013bayesian}. This would require approximating the evidences $f(\pmb y|\boldsymbol{\theta}^{(i)})$ which is possible within EP \cite{minka2013expectation}. However, this approach does not, in general, provide reconstructed images of satisfactory quality as only a few partitions effectively contribute to the final result and blocky artefacts remain present. Similarly, the simple Gaussian mixture $\sum_{i=1}^r \frac{1}r Q^{(i)}(\pmb x)$ also leads to poor final MMSE estimates and large posterior variances. Here, we adopt an alternative strategy which does not require the computation/approximation of model evidences, whereby each approximation $Q^{(i)}(\pmb x)$ is seen as an expert, and the final distribution $Q_f(\pmb x)$ representing $\pmb x$ is built as the Product-of-Experts (PoE) \cite{hinton1999products}, i.e., 
\begin{equation}
   Q_f(\pmb x)  \propto \left[\prod\limits_{i=1}^{r} Q^{(i)} (\pmb x)\right]^{\frac{1}{r}}.
\end{equation}
Each approximation $Q^{(i)}(\pmb x)$ is a multivariate Gaussian distribution with mean and covariance matrix denoted by $\pmb m^{(i)}_*$ and ${\bf \Sigma}^{(i)}_*$, and so is $Q_f(\pmb x)$. Its mean and covariance matrix are given by 
\begin{equation}
{\rm Cov}_{Q_f(\pmb x)}(\pmb x) = \left[\frac{1}{r} \sum\limits_{i=1}^{r} ({\bf \Sigma}^{(i)}_*)^{-1}\right]^{-1}, \quad {\mathbb E}_{Q_f(\pmb x)}[\pmb x] = {\rm Cov}_{Q_f(\pmb x)}(\pmb x) \left[\frac{1}{r} \sum\limits_{i=1}^r ({\bf \Sigma}^{(i)}_*)^{-1}\pmb m^{(i)}_*\right].
\label{Eq: EP final posterior_cov}
\end{equation}
While it is possible to use the individual block-diagonal covariance matrices ${\bf \Sigma}^{(i)}_*$ in \eqref{Eq: EP final posterior_cov}, the resulting marginal variances of $Q_f(\pmb x)$ are often smaller than expected. Instead, we only use the marginal variances of the experts to compute the marginal moments of $Q_f(\pmb x)$ and the final image point estimate considered is ${\mathbb E}_{Q_f(\pmb x)}[\pmb x]$, the mean of $Q_f(\pmb x)$. Although the PoE strategy adopted in this paper differs from more traditional image priors based on overlapping patches, the results presented in Section \ref{Sec: Experimental Results} show that the resulting estimated images are on par with existing patch-based methods and that the estimated posterior variances, despite being approximated, can provide useful information about the posterior uncertainty measures of the image intensities.

\section{EP-EM strategy for hyperparameter estimation}
\label{Sec: EP-EM Strategy for Hyperparameter Estimation}

So far, we have treated $\pmb \theta$ in the prior model  
$f_x(\pmb x|\pmb \theta) = \prod\nolimits_{j=1}^J \sum\nolimits_{k=1}^K \omega_k {\mathcal N}(\pmb x_j;\tilde{\pmb \mu}_k, \widetilde{\bf C}_k)$ as a set of known hyperparameters. In this section, with a fixed $K$ in $\boldsymbol{\theta}=\{\omega_k, \tilde{\pmb \mu}_k, \widetilde{\mathbf C}_k,K\}$, a subset of the parameters in $\pmb \theta$ is pre-trained using external images, while another subset of the parameters is retrieved in an unsupervised manner via an Expectation Maximization (EM) scheme leveraging the Gaussian approximations of the proposed EP algorithms. This is motivated by the fact that training patch-based GMMs in an unsupervised fashion, based on data potentially corrupted by non-Gaussian noise and by a linear degradation operator ${\mathbf H}$ (especially convolution operators), is extremely difficult. On the other hand, it is easy to train patch-based GMMs using many external normalised patches and then only estimate a reduced set of scaling/normalization parameters for an individual image. Without loss of generality, a conventional GMM can be trained using an EM algorithm on a set of external clean image patches \cite{zoran2011learning} (or even noisy images \cite{teodoro2015single,niknejad2015image}). However, such images/patches are often normalised beforehand \cite{zoran2011learning,teodoro2015single,niknejad2015image,teodoro2020block}, e.g., the pixel intensities are scaled to $[0,1]$ and the means of the patches are removed prior to training the GMM. Let $\tilde{\boldsymbol{x}}_j$ represent a zero-mean image patch extracted from an image whose intensity is scaled in $[0,1]$. A conventional GMM can be expressed as
\begin{equation}
f(\tilde{\boldsymbol{x}}_j|\{\omega_k, \pmb \mu_k, {\mathbf C}_k\}_{k=1,\dots,K}) = \sum\limits_{k=1}^K \omega_k{\mathcal N}(\tilde{\boldsymbol{x}}_j;\pmb \mu_k, {\mathbf C}_k),
\label{Eq: conventional_GMM}
\end{equation}
and be characterized by $\{\omega_k, \pmb \mu_k, {\mathbf C}_k\}_{k=1,\dots,K}$. Such a GMM can be trained using a set of patches $\{\tilde{\boldsymbol{x}}_j\}_{j=1,\dots,J}$. When denoising/inpainting arbitrary patches using $\{\omega_k, \pmb \mu_k, {\mathbf C}_k\}_{k=1,\dots,K}$, the mean of each noisy patch can be removed first, the resulting centered patches are then restored and finally the original patches means are added back. In more complex scenarios, this empirical approach has limited interest as adding/removing patch means can complicate the resulting noise statistics. If the GMM is trained without patch mean subtraction, a much larger number of components should be used. Moreover, regardless of whether the patch means are subtracted or not, the trained GMM (in particular its covariance matrices) depends significantly on the scale of the images used for training. While in the Gaussian noise case, it is possible to rescale the noisy observations so that the intensities match those of the training images, this is not possible in the Poisson case as the resulting observation noise would no longer be Poisson distributed. To solve these two problems, we propose to adjust the GMM prior by incorporating and estimating an offset and a scale parameter. More precisely, we express patches of $\pmb x$ as $\pmb x_j = \bar{x}_j\pmb 1+\alpha\tilde{\boldsymbol{x}}_j$, 
where $\bar{x}_j$ denotes the mean of the patch $\pmb x_j$, $\tilde{\boldsymbol{x}}_j$ is a zero-mean patch which is assumed to follow the GMM in \eqref{Eq: conventional_GMM}, and $\alpha>0$ is a scaling factor. It follows that 
\begin{equation}
f(\boldsymbol{x}_j|\bar{x}_j,\alpha,\{\omega_k, \pmb \mu_k, {\mathbf C}_k\}_{k=1,\dots,K}) = \sum\limits_{k=1}^K \omega_k{\mathcal N}(\boldsymbol{x}_j;\bar{x}_j\pmb 1+\alpha \pmb  \mu_k, \alpha^2{\mathbf C}_k).
\end{equation}
The mean $\bar{x}_j$ is unknown in practice. It can be approximated by the mean of the noisy patches, but this is prone to errors in particular if the observed image is blurred, and/or corrupted by high levels of noise. Instead of fixing $\bar{x}_j$, it is assigned a Gaussian prior distribution, and shared across all the patches, i.e., $f(\bar x_j)={\mathcal N}(\bar x_j; m_0, s^2), \forall j=1,\ldots,J.$
This simple prior model allows $\bar x_j$ to be marginalised, leading to the final GMM prior 
\begin{equation}
\begin{aligned}
f_x(\pmb x_j|m_0,s^2,\alpha,\{\omega_k, \pmb \mu_k, {\mathbf C}_k\}_{k=1,\dots,K}) & =  \sum\limits_{k=1}^K \omega_k{\mathcal N}(\pmb x_j; m_0\pmb 1 + \alpha \pmb \mu_k,  s^2 \pmb 1 \pmb 1^T + \alpha^2{\bf C}_k),\\
& = \sum\limits_{k=1}^K \omega_k{\mathcal N}( \pmb x_j; \tilde{\pmb \mu}_k,  \widetilde{\bf C}_k) = f_x(\pmb x_j|\boldsymbol{\theta}),
\label{Eq: f_x_new}
\end{aligned}
\end{equation}
where $\tilde{\pmb \mu}_k=m_0\pmb 1 + \alpha \pmb \mu_k$ and $\widetilde{\bf C}_k=s^2 \pmb 1 \pmb 1^T + \alpha^2{\bf C}_k$. In this work, the parameters $\{\omega_k,\pmb \mu_k,{\bf C}_k, K\}$ of the trained GMM were taken from \cite{zoran2011learning} and these parameters were fixed, while the remaining hyperparameters $\pmb \theta^{\dagger} = \{m_0, s^2, \alpha\}$ are estimated.

A classical approach to hyperparameter estimation is via expectation-maximisation (EM) \cite{moon1996expectation}, whereby $\pmb x$ is treated as a latent vector and $\boldsymbol{\theta}^{\dagger}$ is estimated by maximizing the marginal likelihood $f(\pmb y |\boldsymbol{\theta}^{\dagger})$ or marginal posterior distribution $f(\boldsymbol{\theta}^{\dagger}|\pmb y)$. Unfortunately, a simple EM approach is not possible here since it would require computing expectation w.r.t. the exact posterior $f(\pmb x|\pmb y,\boldsymbol{\theta}^{\dagger})$. However, it is possible to use a variational EM method \cite{celeux2003procedures}, whereby the expectations w.r.t. $f(\pmb x|\pmb y,\boldsymbol{\theta}^{\dagger})$ are replaced by expectations w.r.t an alternative approximating distribution (i.e., the EP approximation here), leading to an EP-EM scheme similar to that used in \cite{kim2006bayesian}. Since $\pmb \theta^{\dagger}= \{m_0, s^2, \alpha\}$ contains only three parameters shared across all the $J$ patches, we do not assign $\pmb \theta^{\dagger}$ a hyperprior, as in practice $f(\pmb y |\boldsymbol{\theta}^{\dagger})$ seen as a function of $\pmb \theta^{\dagger}$ is sufficiently concentrated around its mode.

As mentioned in Section \ref{Sec: Final Estimation using Geometric Average}, one of the main reasons for considering independent experts is that the computational load can be distributed. Although ${\pmb \theta}^{\dagger}$ could be shared across all the experts, here, each expert has its own set of hyperparameters $\pmb \theta^{(\dagger, i)}$ ($i=1,\dots,r$) which are estimated independently. To simplify the notation, in the remainder of this section, the expert indices are omitted. The natural joint likelihood to be considered within EP-EM is $f(\pmb x,\pmb y|{\pmb \theta}^{\dagger})$, but this requires the computation of the expected value of the log-prior $\log(f(\pmb x|\pmb \theta^{\dagger}))$, i.e., the logarithm of a weighted sum of multivariate Gaussian p.d.f.s, which makes the estimation of $\pmb \theta^{\dagger}$ intractable. A more suitable GMM formulation consists of introducing, for each patch, a discrete auxiliary variable $z_j\in \{1,\ldots,K\}$ \cite{bishop2006pattern} and the following hierarchical prior model 
\begin{equation}
f(\pmb x|\pmb z,\pmb \theta^{\dagger})  = \prod\limits_{j=1}^J \prod\limits_{k=1}^K \left[ {\mathcal N}(\pmb x_j;\tilde{\pmb \mu}_k, \widetilde{\bf C}_k)\right]^{\delta(z_{j}-k)}, \quad f(\pmb z) = \prod_{j=1}^J f(z_j),
\label{eq:extended GMM}
\end{equation}
with $\pmb z=[z_1,\ldots,z_J]^T$ and $f(z_j=k)=\omega_k, \forall j\in\{1,\ldots,J\}$. By marginalizing $\pmb z$ in \eqref{eq:extended GMM}, we recover the model \eqref{Eq: proposed GMM patch-based prior}. To simplify the description of the EP methods in Sections \ref{Sec: patch-basedEP_Gaussian_observation} and \ref{Sec: Patch-based EP with non-Gaussian noise}, we discussed approximations of the exact posterior $f(\pmb x|\pmb y, \pmb \theta^{\dagger})$ (Section \ref{Sec: patch-basedEP_Gaussian_observation}) and $f(\pmb x,\pmb u|\pmb y, \pmb \theta^{\dagger})$ (Section \ref{Sec: Patch-based EP with non-Gaussian noise}). However, EP can also be used to approximate the extended posterior distributions
\begin{equation}
\begin{aligned}
f(\pmb x,\pmb z|\pmb y, \pmb \theta^{\dagger}) & \propto  f(\pmb y|\pmb x)f(\pmb x|\pmb z, \pmb \theta^{\dagger})f(\pmb z), \hspace{1.45cm}  \textrm{(Gaussian noise case)},\\
f(\pmb x,\pmb u,\pmb z|\pmb y, \pmb \theta^{\dagger}) & \propto  f(\pmb y|\pmb u)f(\pmb u|\pmb x)f(\pmb x|\pmb z, \pmb \theta^{\dagger})f(\pmb z), \quad \textrm{(Poisson noise case)}.
\label{eq:exact_extended_models}
\end{aligned} 
\end{equation}
This can be achieved easily using the EP factorization used in \cite{hernandez2015expectation}, and it leads to an EP approximation that factorises as $Q^{\dagger}(\pmb x)Q^{\dagger}(\pmb z)$ (resp. $Q^{\dagger}(\pmb x)Q^{\dagger}(\pmb u)Q^{\dagger}(\pmb z)$) in the Gaussian noise case (resp. Poisson noise case). Note that we use the notation $Q^{\dagger}(\cdot)$ to highlight the approximations relate to the extended models in \eqref{eq:exact_extended_models}. Irrespective of the noise model considered, $Q^{\dagger}(\pmb z)$ is chosen to be a product of $J$ independent categorical distributions, one for each element of $\pmb z$, which can take $K$ values. It turns out that, for the Gaussian case, $Q^{\dagger}(\pmb x)$ and $Q^{\dagger}(\pmb z)$ in the product above can be obtained from our EP method which approximates $f(\pmb x|\pmb y, \pmb \theta^{\dagger})$. Indeed, it can be shown that $Q^{\dagger}(\pmb x)=Q(\pmb x)$ and that the probabilities involved in $Q^{\dagger}(\pmb z)$ are the weights computed in \eqref{Eq: patch-based tilted GMM}, i.e., $Q^{\dagger}(z_j=k) = \hat \omega_{j,k}$. Similarly, in the Poisson noise case, we have $Q^{\dagger}(\pmb x)Q^{\dagger}(\pmb u)=Q(\pmb x)Q(\pmb u)$. This can be explained by the fact that EP aims to approximate marginal posteriors in both cases (i.e., $Q(\pmb x)\approx \int f(\pmb x,\pmb u|\pmb y, \pmb \theta^{\dagger})\textrm{d}\pmb u=f(\pmb x|\pmb y,\pmb \theta^{\dagger})$ and $Q(\pmb x)^{\dagger}\approx \sum\nolimits_{\pmb z}\int  f(\pmb x,\pmb u,\pmb z|\pmb y, \pmb \theta^{\dagger})\textrm{d}\pmb u =f(\pmb x|\pmb y, \pmb \theta^{\dagger})$), and since both EP schemes rely on the same factorization, with the same covariance constraints, the marginal approximations are the same. 

Using the extended likelihood $f(\pmb y, \pmb x, \pmb z|\pmb \theta^{\dagger})$ (or $f(\pmb y, \pmb x, \pmb u, \pmb z|\pmb \theta^{\dagger})$ in the Poisson case), the resulting EP-EM reduces to 2 sequential steps. As mentioned in Section \ref{Sec: patch-basedEP_Gaussian_observation}, $Q(\pmb x)$ implicitly depends on $\pmb \theta^{\dagger}$ (and $\pmb y$). While the dependence on $\pmb \theta^{\dagger}$ was not explicit so far to simplify the notation, we now use $Q^{\dagger}(\cdot|\pmb \theta^{\dagger})$ as $\pmb \theta^{\dagger}$ is no longer fixed. The $(t)$-th iteration of EM includes the following $\emph{E-step}$ and $\emph{M-step}$ 

\textbf{\emph{E-step}}: When compute the expected value of the extended likelihood, the only factor that depends on $\pmb \theta^{\dagger}$ is $f(\pmb x|\pmb z,\pmb \theta^{\dagger})$, leading to the cost function
\begin{equation*}
C(\pmb \theta^{\dagger}|\pmb \theta^{\dagger}_{(t-1)})  =  {\mathbb E}_{Q^{\dagger}(\pmb x,\pmb z|\pmb \theta^{\dagger}_{(t-1)})}\left[\log f_x(\pmb x|\pmb z, \pmb \theta^{\dagger})\right] = \sum\limits_{j,k} \hat{\omega}_{j,k} {\mathbb E}_{Q^{\dagger}(\pmb x|\pmb \theta^{\dagger}_{(t-1)})}\left[\log {\mathcal N}(\pmb x_j;\tilde{\pmb \mu}_k, \widetilde{\bf C}_k)\right],
\label{Eq: E_step}
\end{equation*}
where $\pmb \theta^{\dagger}_{(t-1)}$ is the value of $\pmb \theta^{\dagger}$ estimated at  $(t-1)$-th iteration, $\hat{\omega}_{j,k}$ is estimated in \eqref{Eq: patch-based tilted GMM} and $(\tilde{\pmb \mu}_k, \widetilde{\bf C}_k)$ are defined in \eqref{Eq: f_x_new}. Since $Q^{\dagger}(\pmb x|\pmb \theta^{\dagger}_{(t-1)})$ is a Gaussian distribution with mean $\pmb m_*$ and covariance matrix $\pmb \Sigma_*$, the cost function takes the form 
\begin{equation*}
\begin{aligned}
C(\pmb \theta^{\dagger}|\pmb \theta^{\dagger}_{(t-1)})     =  -&\frac{1}{2}\sum\limits_{k,j} \hat \omega_{j,k} \log|s^2 \pmb 1\pmb 1^T + \alpha^2 {\bf C}_k|-\frac{1}{2} \sum\limits_{j,k}  \hat \omega_{j,k} {\rm tr}[(s^2 \pmb 1\pmb 1^T + \alpha^2 {\bf C}_k)^{-1}[{\mathbf \Sigma}_*]_j]+c\\
-&  \frac{1}{2}\sum\limits_{j,k} \hat \omega_{j,k} [([\pmb m_*]_j -m_0 \pmb 1 - \alpha \pmb \mu_k)^T(s^2 \pmb 1\pmb 1^T + \alpha^2 {\bf C}_k)^{-1}([\pmb m_*]_j -m_0 \pmb 1 - \alpha \pmb \mu_k)],
\end{aligned}
\end{equation*}
where $c=-(Jr/2)\log(2\pi)$.
    
\textbf{\emph{M-step}:} Estimate $\pmb \theta^{\dagger}_{(t)}$ by solving  $\pmb \theta^{\dagger}_{(t)} = \underset{\pmb \theta^{\dagger}}{\text{argmax}} ~~C(\pmb \theta^{\dagger}|\pmb \theta^{\dagger}_{(t-1)})$. For a given value of $(s^2,\alpha)$, the optimal value of $m_0$ can be obtained analytically by
\begin{equation}
\hat{m}_0(s^2,\alpha) = \frac{\sum\limits_{j,k} \hat \omega_{j,k} ([\pmb m_*]_j-\alpha \pmb \mu_k)^T(s^2 \pmb 1\pmb 1^T + \alpha^2{\bf C}_k)^{-1}\pmb 1}{\sum\limits_{j,k} \hat \omega_{j,k} \pmb 1^T(s^2 \pmb 1\pmb 1^T + \alpha^2 {\bf C}_k)^{-1}\pmb 1}.
\label{Eq: loss_func_m0_s2_alpha}
\end{equation}
Consequently, it is sufficient to optimize $(s^2,\alpha)$. In practice, instead of an exhaustive search on a 2D grid for $(s^2,\alpha)$, we first fix $\alpha$ and update $s^2$ and then fix $s^2$ and optimize $\alpha$. This procedure is repeated until convergence. The pseudo-code of the final EP-EM algorithm is summarized in Algorithm \ref{alg: algorithm_full}. It is important to mention that estimating only the three parameters $(m_0,s^2,\alpha)$ instead of all the weights, means, and covariance matrices of the $K$-component GMM allows for a more reliable hyperparameter estimation. Since these three parameters are shared by the $K$ components of the GMM, their estimation is usually reliable, irrespective of the number of components in the mixture.

Note that in practice, the image to be reconstructed may not be square or its dimension may not match the size of GMM patch-based prior. It is however possible to apply the same GMM patch-based prior without additional training, by simply marginalising elements (e.g., rows or columns) in \eqref{Eq: f_x_new}, to create GMMs for sub-patches. 

\begin{algorithm}[htbp]
\caption{Full description for the proposed patch-based EP algorithm}
\label{alg: algorithm_full}
\begin{algorithmic}[1]
\Require $\pmb y$, $\bf H$, $\sigma^2$ (if Gaussian i.i.d. noise), conventional trained GMM \{$\omega_k$, $\pmb \mu_k$, ${\bf C}_k, K$\}
\Ensure ${\mathbb E}_{Q_f(\pmb x)}[\pmb x]$, ${\rm Cov}_{Q_f(\pmb x)}(\pmb x)$
\For {$i$ = $1$: $r$}
\While{stopping criterion is not satisfied}
\State \textbullet~{Compute $Q^{(\dagger, i)}(\pmb x|\pmb \theta^{\dagger,i})$ using Algorithm \ref{alg: algorithm_Gaussian_likeli} (for Gaussian observation model) or
\State \quad Algorithm \ref{alg: algorithm_Poisson_likeli} (for Poisson observation model)} 
\State \textbullet~{Estimate $\pmb \theta^{\dagger, i} = \{m_0, s^2, \alpha\}^{\dagger, i}$}
\EndWhile
\State {\textbf{return: }$\pmb m^{(i)}_*$, ${\bf \Sigma}^{(i)}_*$, $\pmb \theta^{\dagger, i}$}
\EndFor
\State Compute final moments via \eqref{Eq: EP final posterior_cov} to obtain ${\rm Cov}_{Q_f(\pmb x)}(\pmb x)$, ${\mathbb E}_{Q_f(\pmb x)}[\pmb x]$
\end{algorithmic}
\end{algorithm}
    
\section{Experiments}
\label{Sec: Experimental Results} In this section, we demonstrate the performance of the proposed patch-based EP algorithm on different image restoration tasks: image denoising, mild image inpainting, and supervised image deconvolution, assuming in each case that the degradation operator ${\mathbf H}$ is known. The experiments are conducted on synthetic grayscale images for which the ground truth image $\pmb x \in {\mathbb R}^{N}$ is known. Two types of operator ${\mathbf H}\in {\mathbb R}^{N\times N}$ are generated.

\begin{itemize}
    \item ${\mathbf H}$ is diagonal: it is either (\romannumeral 1) an identity matrix (for denoising) or (\romannumeral 2) a diagonal mask matrix with diagonal elements equal to 1 if pixels are observed and 0 otherwise (for inpainting). These operators do not induce correlation between pixels and the covariance matrices of the approximating factors $q_{x, 1}(\pmb x)$, $q_{x, 0}(\pmb x)$ are kept diagonal. Although spatial correlations are induced by the patch-based prior, maintaining $q_{x, 1}(\pmb x)$ and $q_{x, 0}(\pmb x)$ as diagonal helps to reduce the computational cost in \eqref{Eq: patch-based tilted GMM} and it is found in practice that the performance is not significantly affected by imposing such constraints for denoising and inpainting. We also demonstrate the benefits of the hyperparameter estimation in these two tasks. 
    \item ${\mathbf H}$ is a 2D convolution matrix, which induces spatial correlation across nearby pixels. In this case, the covariance matrices of the approximating factors $q_{x, 1}(\pmb x)$, $q_{x, 0}(\pmb x)$ are block-diagonal.
\end{itemize}
As mentioned in Section \ref{Sec: EP-EM Strategy for Hyperparameter Estimation}, the GMM used here to build the prior models is taken from \cite{zoran2011learning} (available at \url{https://people.csail.mit.edu/danielzoran/}). It was trained using a set of $2\times 10^6$ patches from a generic external clean dataset \cite{martindatabase} and it consists of $K=200$ components patches of $8\times 8$ pixels. The images restored using EP-EM are obtained via the mean of $Q_f(\pmb x)$, i.e., $\hat {\pmb x}_{\rm proposed} = {\mathbb E}_{Q_f(\pmb x)}[\pmb x]$. To evaluate quantitatively the quality of a restored image $\hat {\pmb x}$, we use the Peak Signal-to-Noise Ratio (PSNR) defined as
\begin{equation}
{\rm PSNR}(\pmb x, \hat{\pmb x}) = 10 \times\log_{10} \frac{(\max (\pmb x))^2}{\frac{1}{N}||\hat {\pmb x}-\pmb x||^2}.
\end{equation}
To visualise the pixel-wise approximate uncertainty associated with $\hat {\pmb x}_{\rm proposed}$, we consider the marginal variances of ${\rm Cov}_{Q_f(\pmb x)}(\pmb x)$. Apart from the denoising/inpainting problems with Gaussian/Poisson noise for which the EP approximation coincides with the actual posterior distribution, it is difficult to assess the reliability of the approximate posterior uncertainties, as the computation of the exact posterior uncertainties is not tractable. To investigate how far actual pixel values are from the estimated region of high posterior density, we compute the pixel-wise intervals covering $95\%$ ($2.5\%-97.5\%$) of the approximate marginal posteriors, centered around the estimated mean. Binary images indicating the pixels for which the ground truth intensity falls into such interval are then produced, and we also report the fraction of pixels in that binary image, for which the ground truth intensities lie in those intervals.

All the experiments have been implemented in MATLAB on an Intel(R) Core(TM) i7-8700K CPU @ 3.70GHz workstation. The proposed EP algorithm was performed without parallel implementation, i.e., the $J$ image patches and $r$ experts are processed using a sequential \textit{for loop}, and execution times are reported for comparison with the competing methods of each scenario.

\subsection{Gaussian i.i.d. noise model}

This subsection demonstrates the effectiveness of the proposed patch-based EP algorithm when the observation noise is Gaussian and i.i.d. with known variance $\sigma^2$. 

\subsubsection{Image denoising and mild inpainting} 

Three benchmark grayscale images (\textit{Cameraman, House, Lena}) of size 
$256 \times 256$ pixels are used as illustrative examples for image denoising and mild inpainting.  The pixel intensities are scaled between $0$ and $1$ so that the same conventional GMM \{$\omega_k$, $\pmb \mu_k$, ${\bf C}_k$, $K$\} can be used within all the competing patch-based methods \cite{zoran2011learning,teodoro2015single}

\begin{figure}[!bp]
\centering
\subfloat[PSNR (dB) comparison.]{\label{fig: PSNRtable_Gauss} \includegraphics[width=0.4\textwidth]{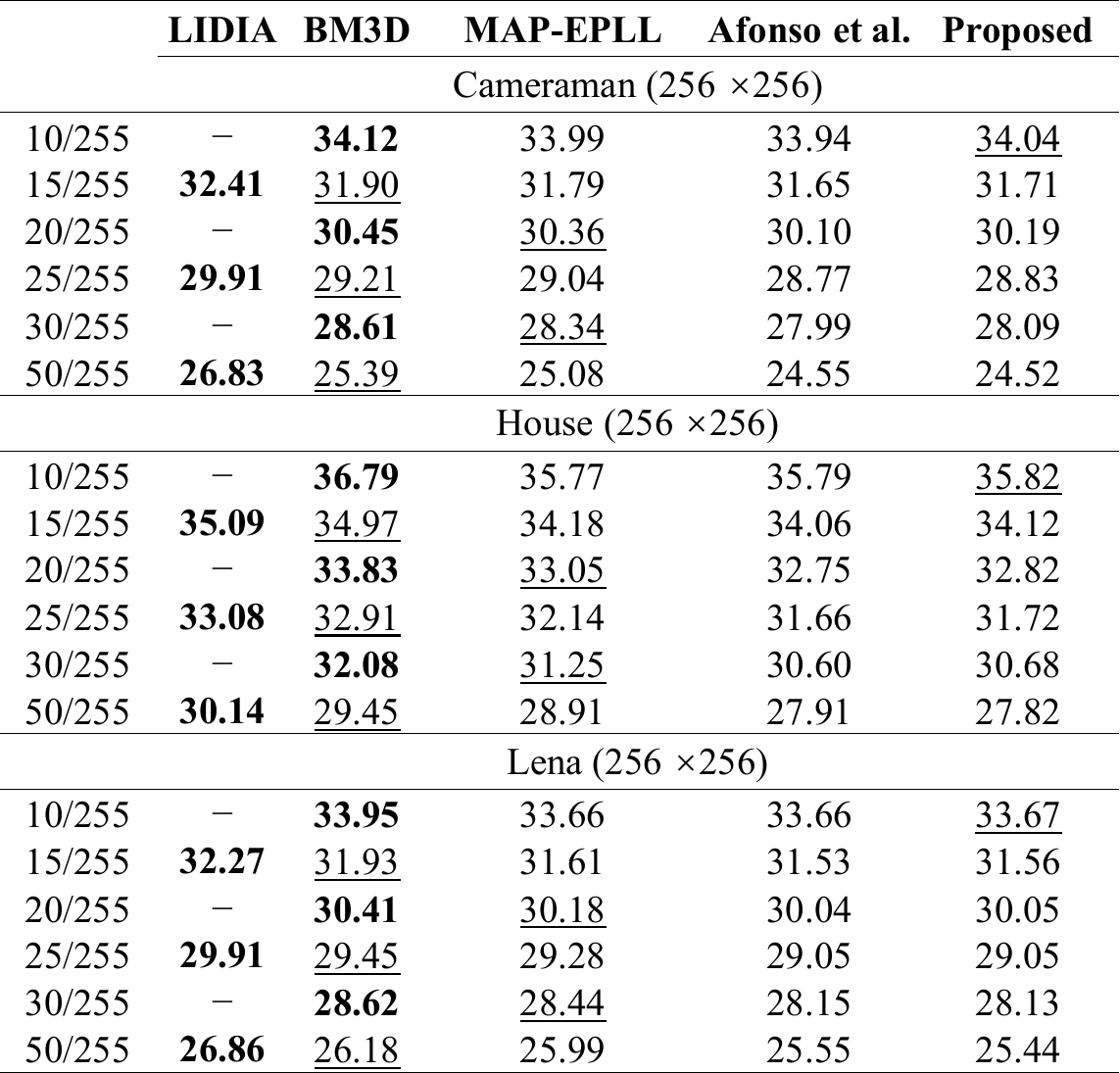}}\quad
\subfloat[Images denoised by the proposed EP-EM algorithm ($\sigma=25/255)$.]{\label{fig: UQmaps_Gauss}\includegraphics[width=0.4\textwidth]{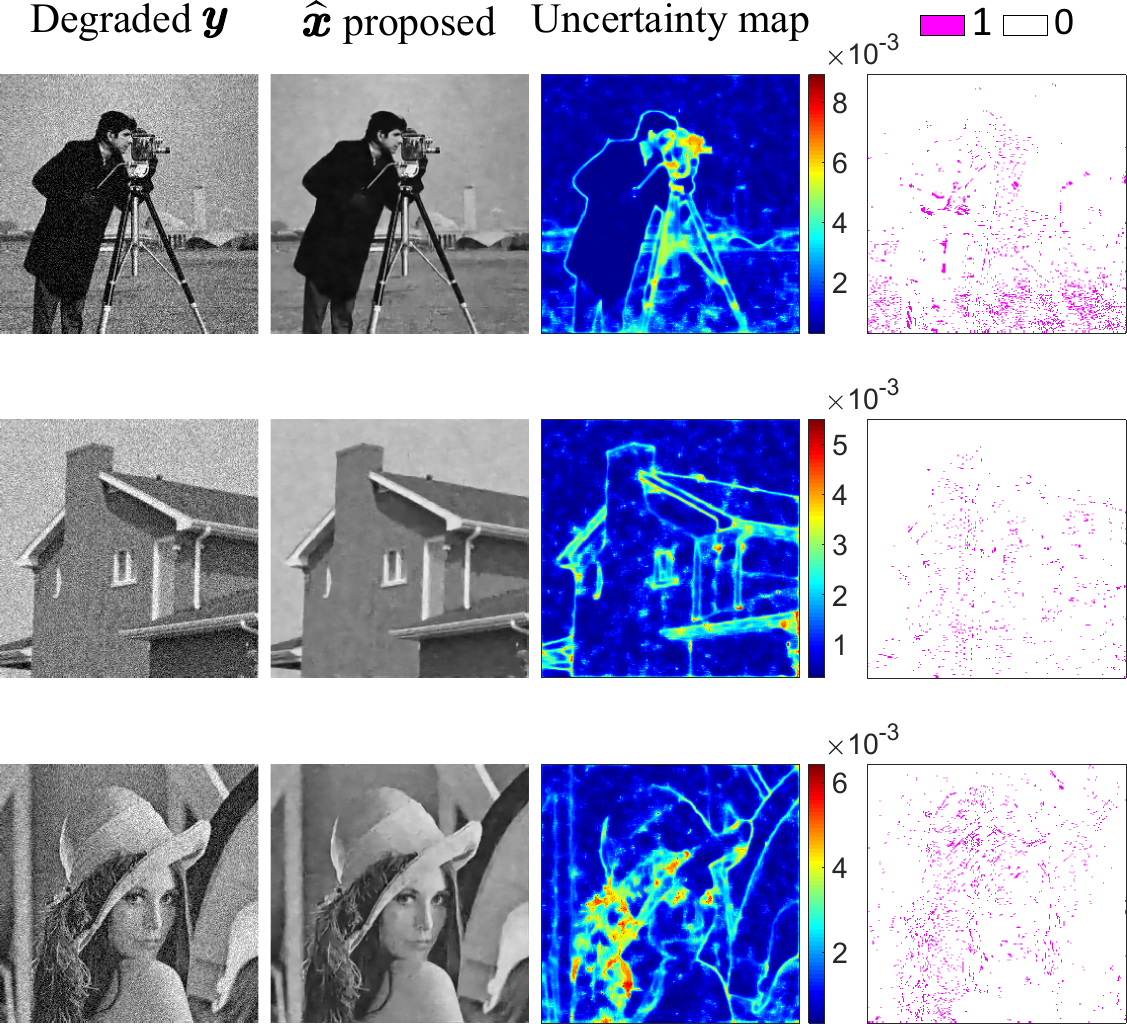}}
\caption{Denoising results for Gaussian i.i.d. noise. (a) PSNR (dB) comparison of different denoising methods. The highest and second highest PSNR values are in bold and underlined, respectively. (b) Restored images obtained by the proposed EP-EM algorithm (second column), associated uncertainty maps (third column), and binary maps indicating pixels for which ground truth intensity falls into $2.5\%-97.5\%$ credible interval (last column). Pixel is depicted in white ('0') if the intensity falls into the credible region and in pink ('1') otherwise.}
\end{figure}

\begin{figure}[!ht]
\centering
\subfloat[Denoising with Gaussian i.i.d. noise ($\sigma$ = 25/255).]{\label{fig: UQ_accuracy_Gauss} \includegraphics[width=0.37\textwidth]{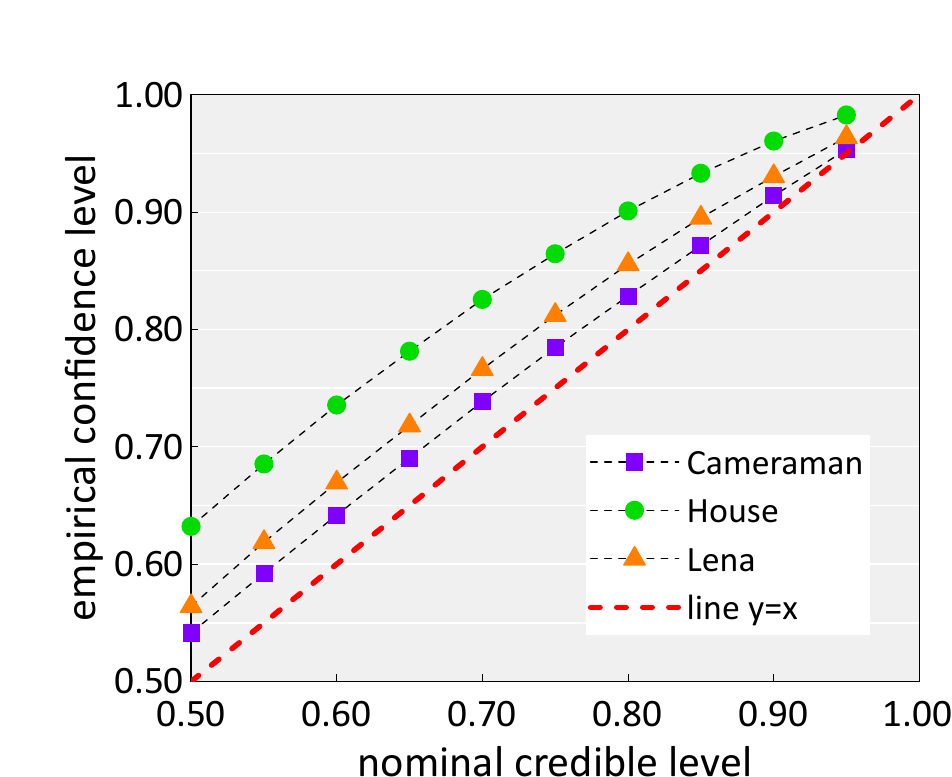}}
\quad
\subfloat[Denoising with Poisson noise (peak value = 30).]{\label{fig: UQ_accuracy_Pois}\includegraphics[width=0.37\textwidth]{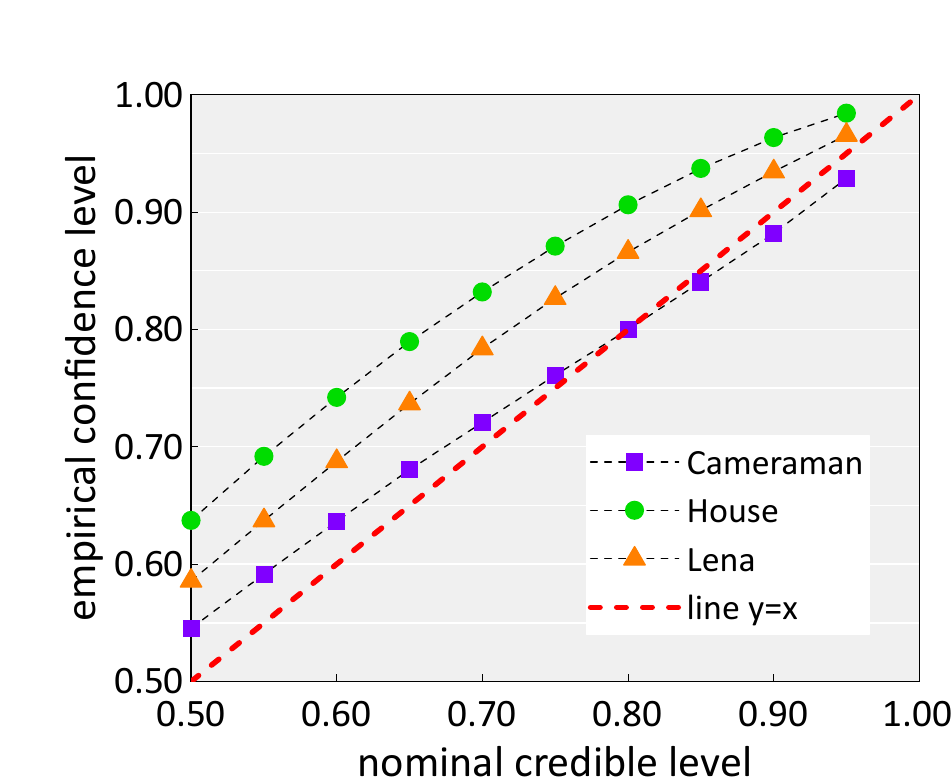}}
\caption{Fraction of pixels inside the different credible intervals for image restored by the proposed EP algorithm.}
\label{fig: UQ_accuracy}
\end{figure}

For the denoising experiments, the noise standard deviation is chosen from the set $\sigma \in \{10,15,20,25,30,50\}/255$, which corresponds to PSNR values about $\{28,25,22,20,19,15\}$dB for the three images. The denoising performance of the proposed EP algorithm is compared to several recent denoising methods assuming Gaussian noise. LIDIA \cite{vaksman2020lidia} is taken as a benchmark to see how the proposed method differs from state-of-the-art deep learning methods, while more conventional popular patch-based methods are also considered, i.e., BM3D \cite{dabov2007image}, the GMM patch-based method proposed in \cite{teodoro2015single} and referred to as \enquote{Afonso et al.}, and the GMM patch-based method using MAP estimation proposed in \cite{zoran2011learning}, referred to as \enquote{MAP-EPLL}. The table in Figure \ref{fig: PSNRtable_Gauss} compares the PSNR values obtained by different methods. It can be observed that the proposed EP algorithm obtains PSNRs slightly worse than BM3D and LIDIA.
Using the same trained GMM as patch-based prior, the proposed EP algorithm provides close PSNRs to MMSE patch-based method \cite{teodoro2015single} and MAP patch-based  method \cite{zoran2011learning}, illustrating its effectiveness as an approximate MMSE denoiser. Recall that the main goal of the proposed EP algorithm is not necessarily to achieve the highest PSNRs (although high PSNRs are preferred), but instead to propose a scalable EP framework with patch-based prior. Still, it can provide comparable PSNRs for denoised images with additional uncertainty estimation. Such example results are presented in Figure \ref{fig: UQmaps_Gauss}. The uncertainty maps exhibit high uncertainty regions around object boundaries and in regions containing complex textures. The far-right column depicts the pixels in pink for which the actual pixel intensity is not included in the corresponding credible interval. The fraction of pixels inside the $2.5\%$-97.5$\%$ credible interval is $95.34\%$ (\textit{Cameraman}), $98.28\%$ (\textit{House}), and $96.38\%$ (\textit{Lena}). While the $95\%$ chosen to define the credible intervals in Figure \ref{fig: UQmaps_Gauss} does not directly relate to these numbers, the result in Figure \ref{fig: UQ_accuracy_Gauss} shows that the fraction of pixels inside the credible intervals drops nearly linearly as the width of the credible intervals shrinks, which indicates that the proposed EP method does not, on average across all the pixels, drastically underestimate or overestimate the marginal uncertainties in these cases. 

\begin{figure}[t]
\centerline{\includegraphics[width=0.75\textwidth]{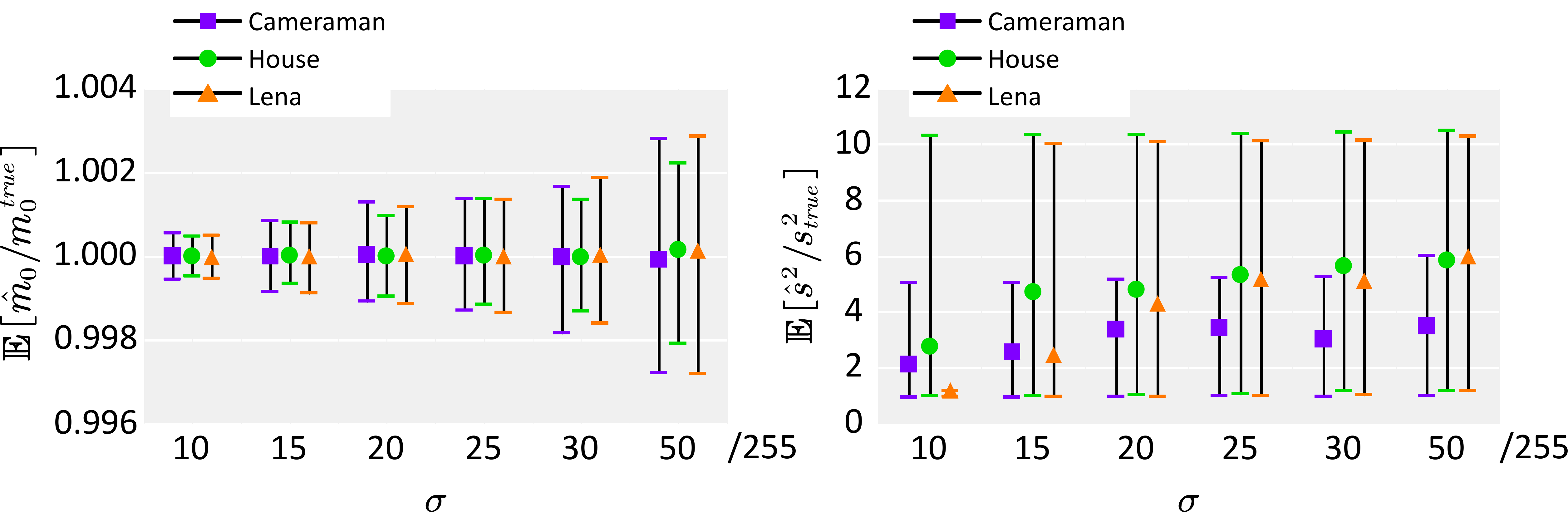}}
\caption{Results of hyperparameter estimation in denoising for Gaussian i.i.d. noise. $\hat m_0$, $\hat s^2$ are the estimates obtained by the proposed EP-EM method. Y-axis $\mathbb{E}[\hat m/m^{true}]$ and $\mathbb{E}[\hat s^2/s_{true}^2]$ are the $90 \%$ confidence intervals of the expected value of the estimated values divided by the ground truth values of three test images.}
\label{fig: result_hyperparameter_estimation_Gaussian}
\end{figure}

The hyperparameters $\{m_0, s^2\}$ have been estimated by EP-EM while running the experiments presented in the table of Figure \ref{fig: PSNRtable_Gauss}. Since the image intensity was already scaled in $[0,1]$, the scaling factor $\alpha$ has been fixed to 1 here. For completeness, EP-EM has been tested using the three images with $200$ noise realizations for each standard deviation $\sigma$, resulting in $200$ estimates $(\hat m_0,\hat s^2)$ per expert. In practice, the results obtained by different experts are similar and we only present the results of one expert, selected randomly. The average ratios $\hat m_0/{m_0^{true}}$ and $\hat s^2/{s^2_{true}}$ and the associated 5th-95th quantile intervals, are displayed in Figure \ref{fig: result_hyperparameter_estimation_Gaussian}. It can be seen that the $90\%$ confidence intervals of $\mathbb{E}[\hat m_0/{m_0^{true}}]$ and $\mathbb{E}[\hat s^2/{s^2_{true}}]$ become larger as the noise variance increases, indicating that less information become available. However, all the experts provide similar results as they all use the same data. Consequently, although in Algorithm \ref{alg: algorithm_full} we suggest using EP-EM for each expert, if communication is possible between experts, a single expert can perform the hyperparameter estimation and pass on its final estimates to the others, which can then only use Algorithm \ref{alg: algorithm_Gaussian_likeli} for image restoration.

For denoising tasks with images of size $256\times 256$ pixels, the execution time of Afonso et al.'s method when taking non-overlapping patches is about 0.3 seconds, while each expert in the proposed EP-EM algorithm without parallel implementation takes 10.8 seconds on average (the algorithm converges in 2 to 3 iterations, and each iteration takes about 5 seconds). The costly computation is the grid search in finding the optimized parameters of $m_0$ and $s^2$. For only EP with fixed hyperparameters, it takes a similar execution time as the method of Afonso et al. LIDIA takes about 9 seconds, BM3D takes about 1.6 seconds, and MAP-EPLL takes about 19.1 seconds. 

To assess the performance of the methods for inpainting problems, observed images $\pmb y$ are simulated via $\pmb y \sim {\mathcal N}(\pmb y; {\mathbf H}\pmb x, \sigma^2 {\mathbf I})$, where ${\mathbf H}$ is a diagonal matrix with $40\%$, $60\%$, $80\%$ diagonal elements randomly being zero, respectively, and $\sigma$ = \{10, 20, 30\}/255. Table \ref{Table: comparison_result_inpainting_Gaussian} reports the PSNRs of the inpainting results obtained using the proposed algorithm and two competing methods in \cite{zoran2011learning} and \cite{teodoro2015single}. It can be seen that here, the proposed EP-EM algorithm provides higher PSNR values than the MMSE patch-based GMM method \cite{teodoro2015single} and MAP patch-based GMM method \cite{zoran2011learning}. 

For these inpainting experiments, MAP-EPLL takes about 376.3 seconds, the method of Afonso et al. with only non-overlapping patches takes about 34.6 seconds, while a single expert of the proposed EP-EM method takes about 204.5 seconds on average (the algorithm converges in 7 iterations and each iteration takes about 33.1 seconds). The hyperparameter estimation takes about 4.5 seconds at each iteration. 

\begin{table}[t]
\caption{Results of joint inpainting and denoising with Gaussian i.i.d. Noise. PSNR (dB) comparison of different methods. The fraction of missing pixels is $\{40\%, 60\%, 80\%\}$, with noise standard deviation $\sigma = \{10,20,30\}/255$.}
\centering
\resizebox{0.7\textwidth}{!}{
\begin{tabular}{l|c|c|c|c|c|c|c|c|c|l}
\toprule
\multirow{2}{*}{} & \multirow{2}{*}{$\sigma$} & \multicolumn{3}{c|}{Cameraman (256$\times$256)} & \multicolumn{3}{c|}{House (256$\times$256)} & \multicolumn{3}{c}{Lena (256$\times$256)} \\ \cline{3-11} 
& 
&\multicolumn{1}{c}{40$\%$}&
\multicolumn{1}{c}{60$\%$}& \multicolumn{1}{c|}{80$\%$}& \multicolumn{1}{c}{40$\%$}& \multicolumn{1}{c}{60$\%$}& 
\multicolumn{1}{c|}{80$\%$}& 
\multicolumn{1}{c}{40$\%$} & 
\multicolumn{1}{c}{60$\%$} & 
\multicolumn{1}{c}{80$\%$} \\ 
\hline
Afonso et al. & \multirow{3}{*}{$10/255$} &  \multicolumn{1}{c}{29.11} &  \multicolumn{1}{c}{26.94} & \multicolumn{1}{c|}{23.90} & \multicolumn{1}{c}{33.65} & \multicolumn{1}{c}{32.04} &  \multicolumn{1}{c|}{29.06} &
\multicolumn{1}{c}{30.17} &  \multicolumn{1}{c}{28.08} & 
\multicolumn{1}{c}{25.56}   \\ 
MAP-EPLL & &
\multicolumn{1}{c}{29.05} &
\multicolumn{1}{c}{27.14} & 
\multicolumn{1}{c|}{24.13} &
\multicolumn{1}{c}{33.70} & 
\multicolumn{1}{c}{31.97} &
\multicolumn{1}{c|}{28.61} & 
\multicolumn{1}{c}{29.93} &
\multicolumn{1}{c}{27.80} & 
\multicolumn{1}{c}{25.23}   \\  
Proposed &   & 
\multicolumn{1}{c}{\textbf{29.74}}  &
\multicolumn{1}{c}{\textbf{27.56}}  &
\multicolumn{1}{c|}{\textbf{24.24}} &
\multicolumn{1}{c}{\textbf{34.21}}  &
\multicolumn{1}{c}{\textbf{32.75}}  &
\multicolumn{1}{c|}{\textbf{29.38}}  &
\multicolumn{1}{c}{\textbf{30.53}}  &
\multicolumn{1}{c}{\textbf{28.36}}  & 
\multicolumn{1}{c}{\textbf{25.84}}   \\ 
\hline
Afonso et al. & \multirow{3}{*}{$20/255$} & \multicolumn{1}{c}{27.48}    & \multicolumn{1}{c}{25.75}   & 23.08   &   \multicolumn{1}{c}{31.07}  &  \multicolumn{1}{c}{29.68}  &   27.03 &\multicolumn{1}{c}{28.01}    &   \multicolumn{1}{c}{26.64} & \multicolumn{1}{c}{24.63}   \\
MAP-EPLL &    &  \multicolumn{1}{c}{26.74}   &  \multicolumn{1}{c}{25.05}   &  22.28  &\multicolumn{1}{c}{30.27}    &\multicolumn{1}{c}{28.96}    &  \multicolumn{1}{c|}{25.49}  &\multicolumn{1}{c}{26.57}    & \multicolumn{1}{c}{25.21} 
& \multicolumn{1}{c}{23.03}   \\  
Proposed &   &\multicolumn{1}{c}{\textbf{27.85}}    & \multicolumn{1}{c}{\textbf{26.11}}   & \multicolumn{1}{c|} {\textbf{23.41}}   &\multicolumn{1}{c}{\textbf{31.36}}   & \multicolumn{1}{c}{\textbf{30.05}} & \multicolumn{1}{c|}{\textbf{27.15}}   & \multicolumn{1}{c}{\textbf{28.15}}  &\multicolumn{1}{c}{\textbf{26.81}}   &  \multicolumn{1}{c}{\textbf{24.77}}  \\ 
\hline
Afonso et al. & \multirow{3}{*}{$30/255$} & \multicolumn{1}{c}{26.04}   &  \multicolumn{1}{c}{24.45}  & 22.30   &\multicolumn{1}{c}{29.22}   &\multicolumn{1}{c}{27.56}    & 25.34   &\multicolumn{1}{c}{26.55}    & \multicolumn{1}{c}{25.22}   &\textbf{23.49}    \\ 
MAP-EPLL & &\multicolumn{1}{c}{24.72}    & \multicolumn{1}{c}{23.09}  & 20.66    & \multicolumn{1}{c}{28.24}   &\multicolumn{1}{c}{31.88}    & \multicolumn{1}{c|}{23.18}   &  \multicolumn{1}{c}{24.72}  & \multicolumn{1}{c}{23.50}   & 21.02   \\  
Proposed &                   &  \multicolumn{1}{c}{\textbf{26.24}} & \multicolumn{1}{c}{\textbf{24.60}}  & \multicolumn{1}{c|}{\textbf{22.59}}    & \multicolumn{1}{c}{\textbf{29.26}}   &\multicolumn{1}{c}{\textbf{27.61}}    &\multicolumn{1}{c|}{\textbf{25.37}}    &\multicolumn{1}{c}{\textbf{26.57}}    & \multicolumn{1}{c}{\textbf{25.33}}   &  \multicolumn{1}{c}{\textbf{23.49}}  \\ 
\bottomrule 
\end{tabular}}
\label{Table: comparison_result_inpainting_Gaussian}
\end{table}

\subsubsection{Deconvolution}

\begin{figure}[!ht]
\centerline{\includegraphics[width=0.8\textwidth]{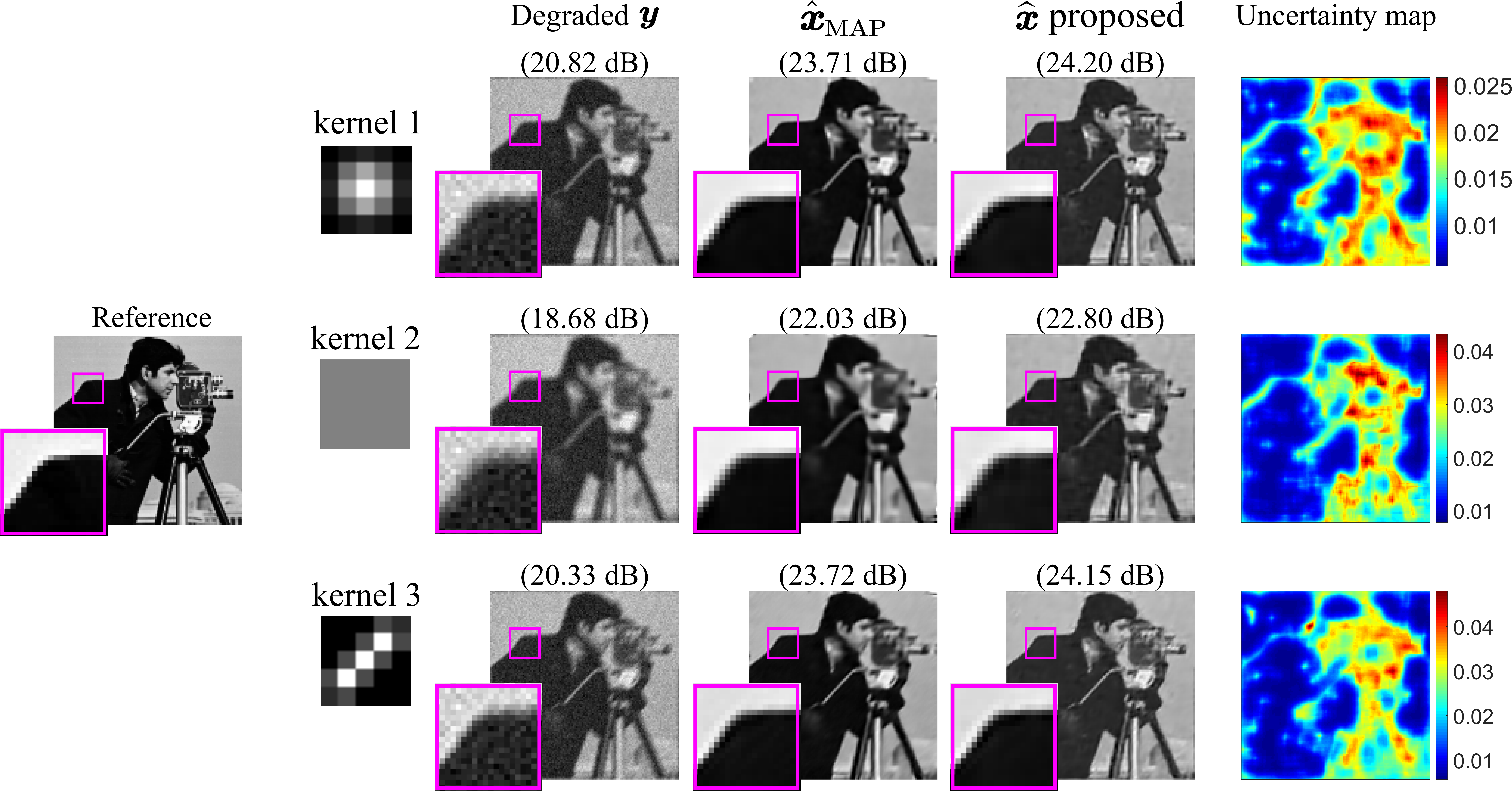}}
\caption{Results for deconvolution with Gaussian i.i.d. noise, $\sigma=0.05$. First column: reference of ground truth image and a zoomed-in region. Second column: three kernels used in simulation. The size of the three kernels are both $5\times 5$ pixels.  Kernel 2 is a uniform kernel. ${\mathbf H}$ is non-invertible constructed by kernel 1, and invertible constructed by kernels 2 and 3. Third column: degraded observations with different kernels. Fourth column: estimated images by MAP-EPLL \cite{zoran2011learning}. Fifth column: estimated images by the proposed EP-EM method. Last column: uncertainty maps for the estimated images obtained by the proposed method.}
\label{fig: result_GaussianDeconv_posmean}
\end{figure}

For deconvolution experiments, a sub-image of $128\times 128$ pixels taken from the $\textit{Cameraman}$ image is used as ground truth $\pmb x$, and convolved by three different kernels of size $ 5\times 5$ pixels, as shown in Figure \ref{fig: result_GaussianDeconv_posmean} (i.e., a Gaussian blur, a uniform blur and a motion blur). The proposed EP-EM method is compared to the MAP estimator MAP-EPLL \cite{zoran2011learning}. It can be seen that the PSNR values after deconvolution using the proposed EP-EM method are higher than the MAP estimator. The zoom-in boxes highlight the boundary between distinct homogeneous regions. The four uncertainty maps in the last column show that the pixels with high uncertainty generally appear at the edges of objects, while subtle changes can be observed when using different blur kernels. 

For deconvolution of images of size $128\times 128$ pixels, the execution time of MAP-EPLL is about 7.0 seconds, while one expert of the proposed EP-EM methods takes about 750.1 seconds for 20 iterations (each iteration takes about 38.0 seconds and no parallel implementation was used. 20 samples are generated in RBMC to approximate ${\rm Cov}_{P_1}(\pmb x)$ in \eqref{Eq: titled_q1_Gaussian}). 

\subsubsection{Comparison with EP with full covariance matrices} This subsection assesses the performance degradation of both the approximate MMSE estimates and the posterior uncertainty measures obtained by the proposed EP algorithm (Algorithm \ref{alg: algorithm_Gaussian_likeli}), when compared with more traditional EP implementation (referred to as naive EP), where all the approximating factors have full covariance matrices.

\begin{figure}[t]
\centering
\subfloat[Approximate MMSE and associated uncertainty maps obtained by naive EP and the proposed EP algorithms.]{\label{fig: comparison_naifEP_1 } \includegraphics[width=0.35\textwidth]{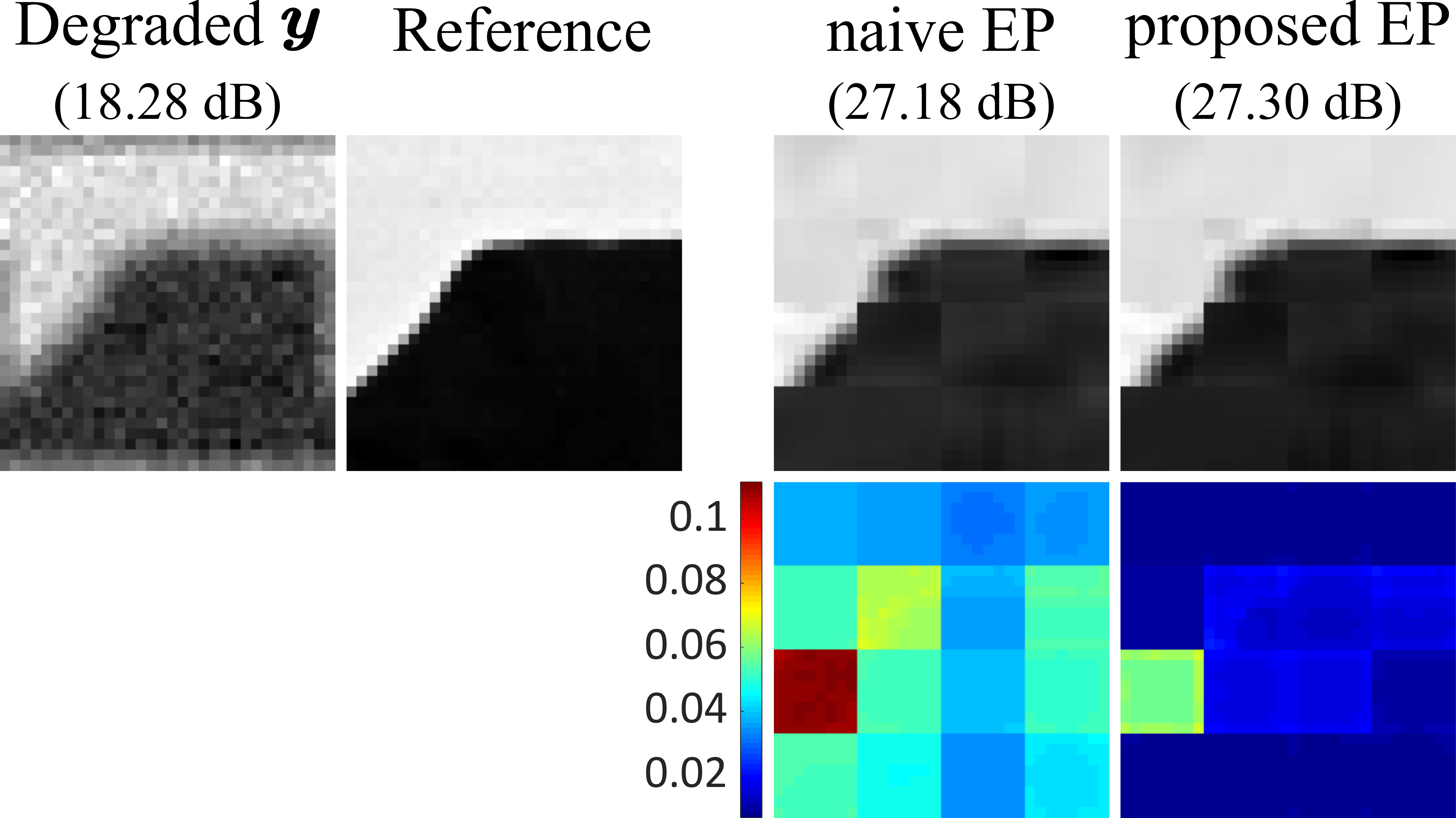}}\quad \quad 
\subfloat[Marginal posterior variances obtained by naive EP and the proposed EP algorithms.]{\label{fig: comparison_naifEP_2 }\includegraphics[width=0.35\textwidth]{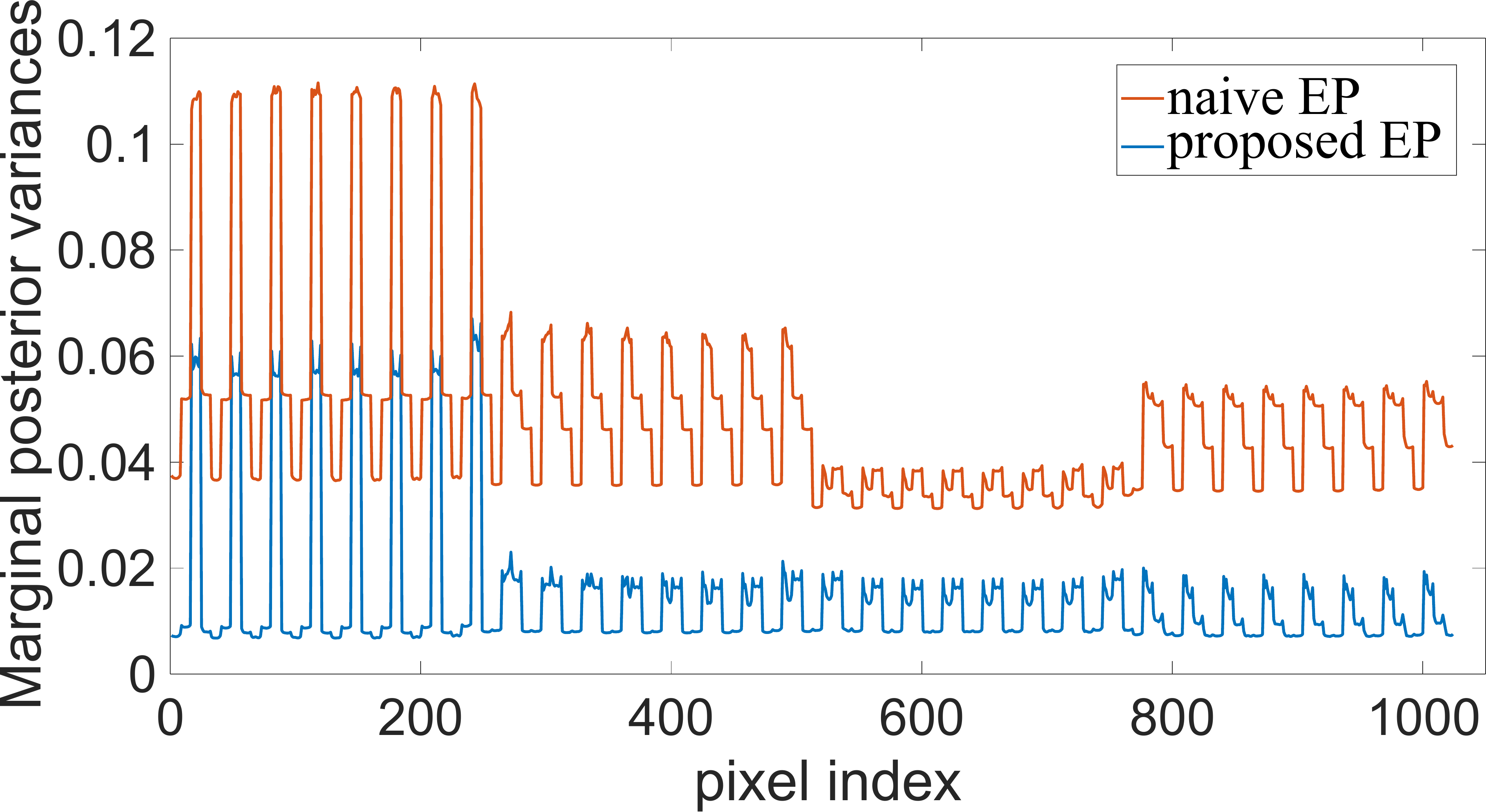}}
\caption{Comparison of approximate MMSE and posterior uncertainty quantification by naive EP and the proposed EP algorithms.}
\label{fig: comparison_naifEP}
\end{figure}

For image denoising and mild inpainting problems (with Gaussian noise), as discussed earlier, the exact posterior distribution for each expert is tractable and reduces to patch-wise independent GMMs. Thus our proposed EP approximation and the naive implementation both provide marginal means and variances that match those of the exact posterior distribution. 
For the deconvolution problem, the marginal moments estimated by EP are no longer exact. Within the naive implementation, all the covariance matrices of the approximating factors are full and the GMM patch-based prior is approximated by the product of $J$ Gaussian approximating factors, where $J$ is the number of non-overlapping patches in the image. This leads to $(J+1)$ factors that need to be updated sequentially in each EP iteration. To keep the computational cost of the naive implementation relatively low, we only consider a $32 \times 32$ pixels image (a small subset of \textit{Cameraman}), with ${\mathbf H}$ corresponding to a $5\times 5$ pixels uniform kernel. Moreover, we only use a single expert in this comparison, to better visualize the impact on a given posterior distribution. 
The results in Figure \ref{fig: comparison_naifEP} show that the approximate MMSE estimates obtained by the naive EP and the proposed EP methods are comparable. In this example, it seems that the marginal variances of the naive EP method are larger than those of the proposed EP method. While we think that this is due to the fact that the block-diagonal approximation neglects part of the actual correlations, it could also be that the naive EP implementation overestimates uncertainties due to convergence issues during the gradient-based iterative updates used to ensure that all the approximate covariance matrices are strictly positive-definite. Indeed, the gradient descent used to minimize \eqref{Eq: loss_func_Omega0} might stop when moving in the gradient direction is no longer possible due to the constraints, but it might be possible to further optimize the cost function by moving along another direction (not investigated in this work). Note that the blocky artefacts in Figure \ref{fig: comparison_naifEP} are due to the fact that a only single expert is considered here. In terms of computational cost, the proposed EP algorithm only takes about 1.6 minutes for 20 iterations, while the naive EP method takes about  2  hours  to  obtain  the  posterior  approximations  (for  the  same  20  EP  iterations). In addition, the naive EP can not scale to large images as it requires the inversion of large $N\times N$ covariance matrices, while the proposed EP algorithm allows for greater scalability, inverting only small $r \times r$ covariance matrices.

\subsection{Poisson noise model} This subsection demonstrates the effectiveness of the proposed patch-based EP algorithm when the observation noise being Poisson distributed. The proposed method is compared with VST-BM3D \cite{makitalo2010optimal}, SPIRAL-TV \cite{harmany2011spiral}, and PIDAL-TV \cite{figueiredo2010restoration}. VST-BM3D combines the Anscombe transform \cite{anscombe1948transformation} with the BM3D algorithm. The regularization parameters of SPIRAL-TV and PIDAL-TV are tuned to find the highest PSNRs.

\subsubsection{Denoising and mild inpainting}
\begin{figure}[t]
\centering
\subfloat[PSNR (dB) comparison.]{\label{fig: PSNRtable_Pois} \includegraphics[width=0.4\textwidth]{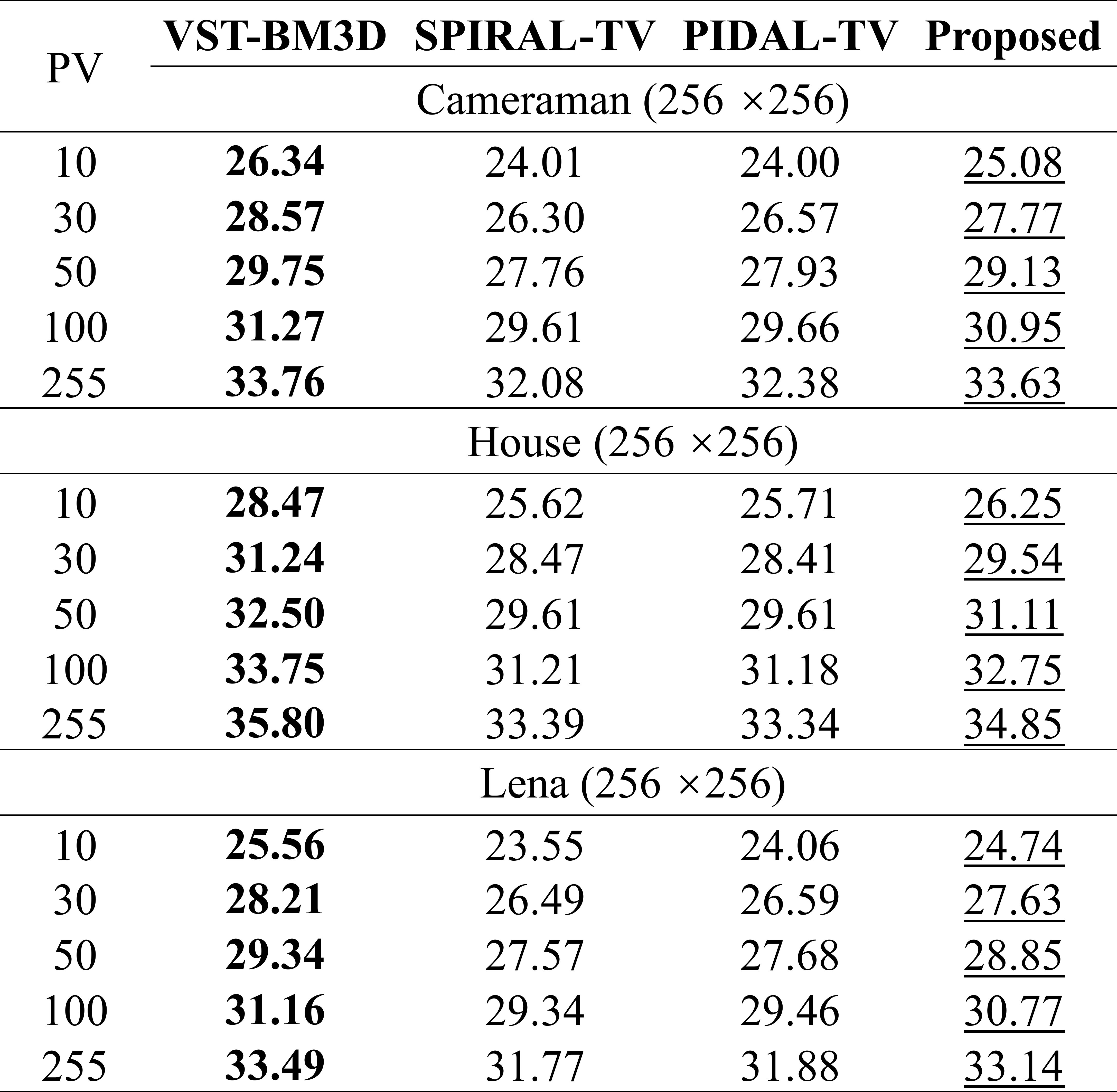}}\quad
\subfloat[Images denoised by the proposed EP-EM algorithm (peak value = 30).]{\label{fig: UQmaps_Pois}\includegraphics[width=0.46\textwidth]{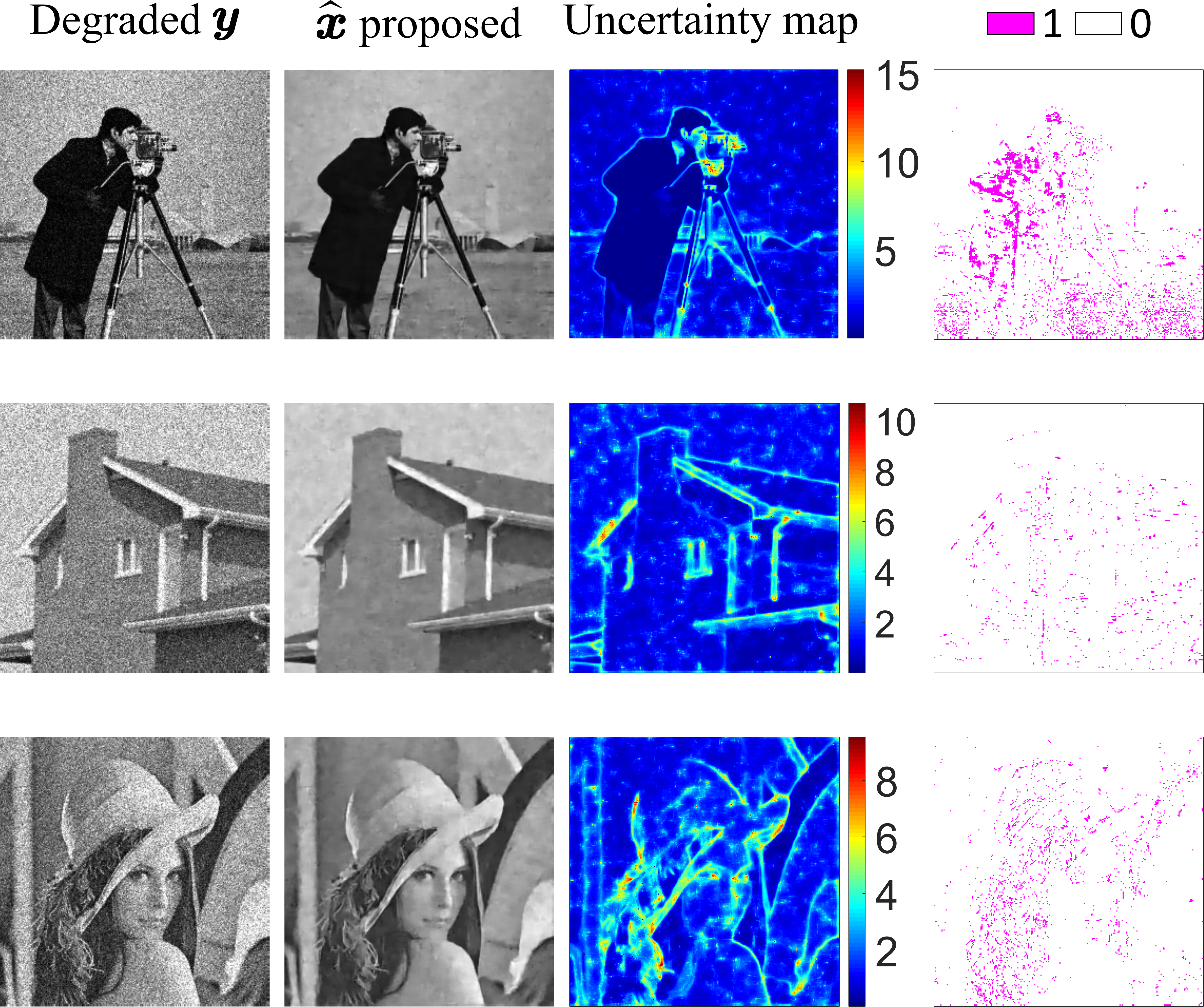}}
\caption{Denoising results for Poisson noise. (a) PSNR (dB) comparison of different denoising methods. The highest and second highest PSNR values are in bold and underlined, respectively. (b) Restored images obtained by the proposed EP-EM (second column), associated uncertainty maps (third column), and binary maps indicating pixels for which ground truth intensity falls into $2.5\%$-$97.5\%$ credible interval (last column). 
Pixel is depicted in white ('0') if the intensity falls into the credible region and in pink ('1') otherwise.}
\label{fig: PSNR_PoissonDenoising}
\end{figure}

The same benchmark grayscale images  (\textit{Cameraman, House, Lena}) as in the previous experiments are used in this section. For the denoising experiments, the observed images $\pmb y$ are simulated using $\pmb y \sim {\mathcal P}(\pmb x)$ and the peak value of $\pmb x$ is set to be in the set of $\{10,30,50,100,255\}$. The table in Figure \ref{fig: PSNRtable_Pois} presents the PSNR values obtained by different denoising methods. VST-BM3D provides higher PSNRs than the proposed EP algorithm, in particular when the peak value is low. This is mainly because in low peak value regime, approximating the Poisson distribution of observations using a Gaussian distribution is difficult and the accuracy loss may be substantial. Compared to SPIRAL-TV and PIDAL-TV, which are non patch-based methods, the proposed EP algorithm performs better and it does not require as much supervision in hyperparameter tunning as the two methods, since a subset of parameters in GMM patch-based prior is pre-trained, and another subset of hyperparameters is retrieved in an unsupervised manner. Similar to denoising experiments for Gaussian i.i.d. noise,  we emphasize that the main goal of the proposed EP algorithm is not to acheive the highest PSRNs. Rather, it aims to be scalable to high-dimensional imaging problems to provide comparable denoising results with additional uncertainty estimation. Figure \ref{fig: UQmaps_Pois} shows the uncertainty maps and binary images associated with uncertainty quantification for peak value of $30$. The fraction of pixels fall into $2.5\%-97.5\%$ credible interval is $92.89\%$ (\textit{Cameraman}), $98.43\%$ (\textit{House}), and $96.57\%$ (\textit{Lena}). Figure \ref{fig: UQ_accuracy_Pois} presents the fraction of pixels inside different credible intervals, which follows the same linear trend as in the denoising experiments with Gaussian noise. 

\begin{figure}[t]
\centerline{\includegraphics[width=0.95\textwidth]{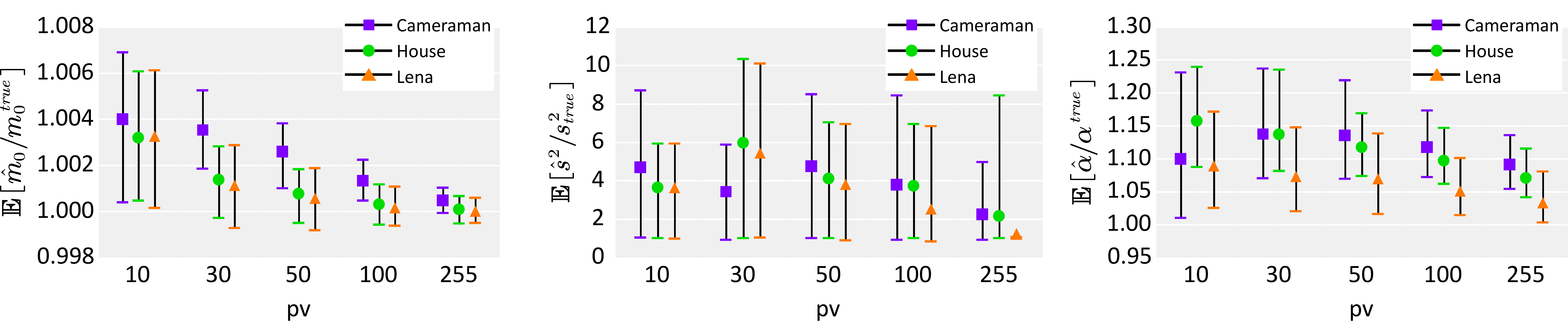}}
\caption{Results for hyperparameter estimation in denoising for Poisson noise. $\hat m_0$, $\hat s^2$, $\hat \alpha$ are the estimates obtained by the proposed EP-EM method. Y-axis $\mathbb{E}[\hat m/m^{true}]$, $\mathbb{E}[\hat s^2/s_{true}^2]$, $\mathbb{E}[\hat \alpha^2/\alpha^{true}]$ are the $90 \%$ confidence intervals of the expected value of the estimated values divided by the ground truth values of three test images.}
\label{Eq: result_hyperparameter_estimation_Poisson}
\end{figure}

The hyperparameter estimation results for the three images are presented in Figure \ref{Eq: result_hyperparameter_estimation_Poisson}.  This figure shows that $90\%$ confidence intervals of the three expected values $\mathbb{E}[\hat m_0/{m_0^{true}}]$, $\mathbb{E}[\hat s^2/{s^2_{true}}]$, $\mathbb{E}[\hat \alpha/{\alpha_{true}}]$ are getting narrower as the peak value increases, indicating that as the peak value of the Poisson observations increases, it becomes easier to estimate the hyperparameters, as the signal-to-noise ratio increases.

\begin{figure}[t]
\centering
\subfloat[Denoising for Gaussian i.i.d. noise ($\sigma=25/255$).]{\label{fig: result_Poisson_vs_Gaussian_denoising_Gauss} \includegraphics[width=0.6\textwidth]{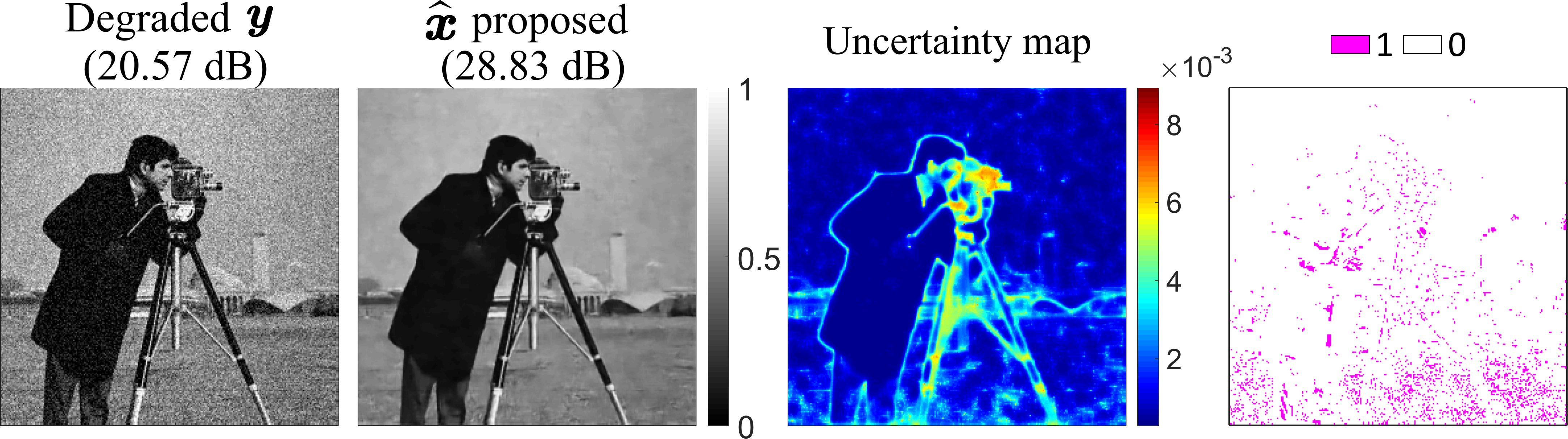}}

\subfloat[Denoising for Poisson noise (peak value =30).]{\label{fig: result_Poisson_vs_Gaussian_denoising_Pois}\includegraphics[width=0.6\textwidth]{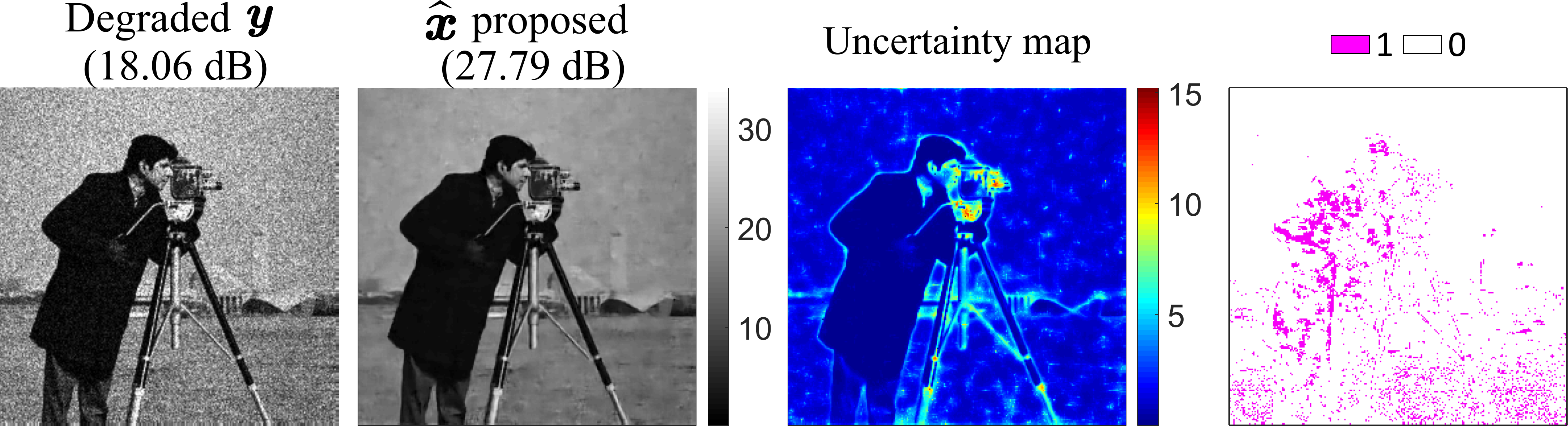}}
\caption{Comparison results between Gaussian and Poisson denoising. The fraction of pixels in $2.5\%$-$97.5\%$ credible interval is (a) 95.34$\%$ and (b) 92.84$\%$.}
\label{fig: result_Poisson_vs_Gaussian_denoising}
\end{figure}

Figure \ref{fig: result_Poisson_vs_Gaussian_denoising} depicts a comparison of denoised \textit{Cameraman} images corrupted by Gaussian i.i.d. noise (Figure \ref{fig: result_Poisson_vs_Gaussian_denoising_Gauss}) and Poisson noise (Figure \ref{fig: result_Poisson_vs_Gaussian_denoising_Pois}) using the proposed EP-EM algorithm. It can be seen that the uncertainty maps present similar yet different features. Both maps exhibit lower uncertainties in homogeneous regions, however the uncertainties in the bottom image (Poisson noise) are lower in homogeneous low-intensity regions (e.g., the \textit{Cameraman}'s coat and hat than in the sky). This is mainly due to two reasons: the Poisson likelihood which induces uncertainties depending on the mean signal values, and the Poisson rectified model which tends to keep most of the mass of $Q(\pmb x)$ in the positive orthant.

\begin{figure}[t]
\centerline{\includegraphics[width=0.6\textwidth]{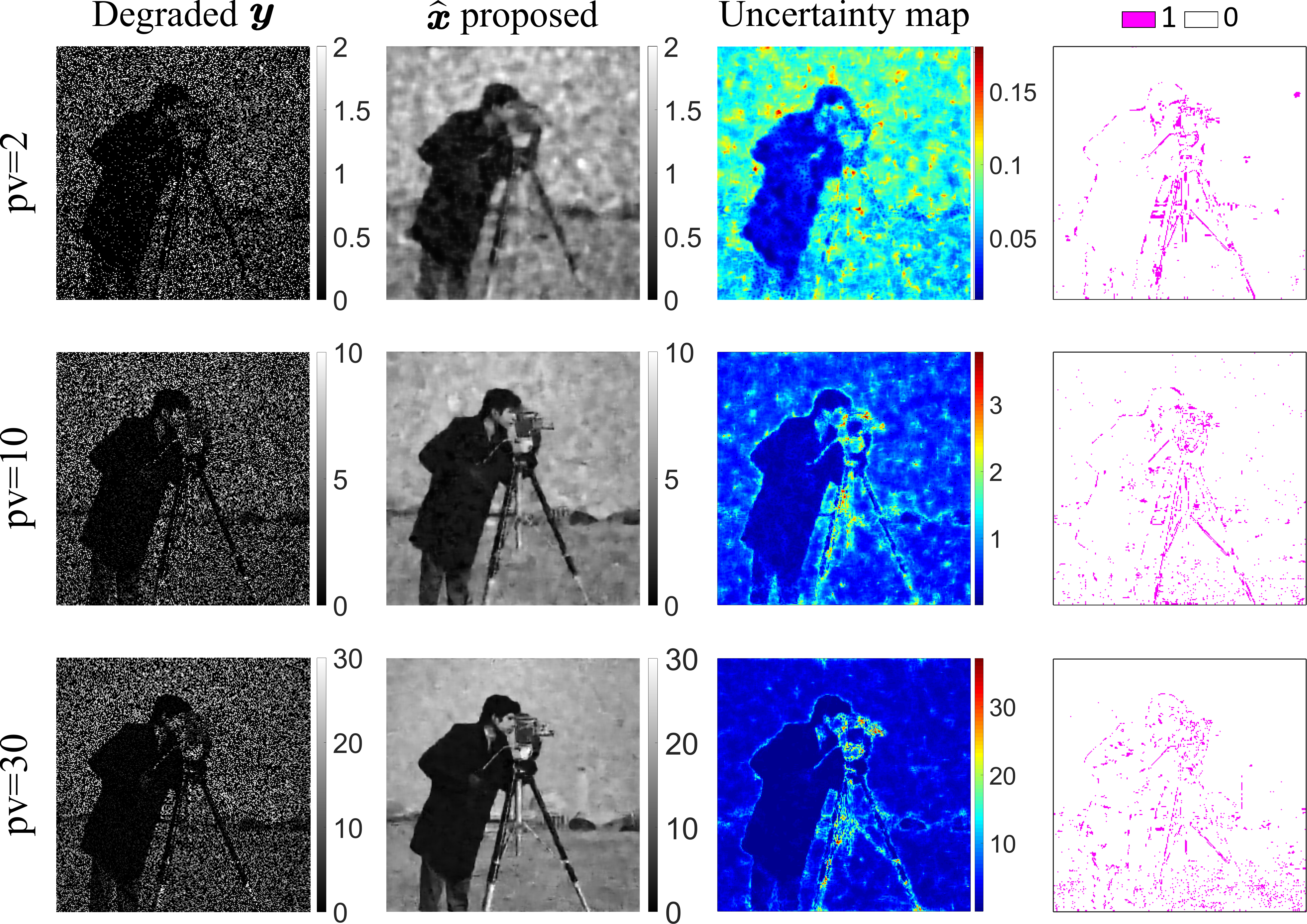}}
\caption{Comparison results of inpainting and denoising with Poisson noise. The percentage of missing pixels is 60 $\%$. PSNR of the estimated images and the fraction of pixels fall into $2.5\%$-$97.5\%$ credible interval are 19.11 dB, 96.71$\%$ (peak value = 2), 22.14 dB, 96.51$\%$ (peak value = 10), and 24.26 dB, 96.25$\%$ (peak value = 30), respectively.}
\label{fig: result_Poisson_inpainting}
\end{figure}

We also conducted inpainting experiments, where the observed images $\pmb y$ are simulated using $\pmb y \sim {\mathcal P}({\mathbf H}\pmb x)$, the degradation operator ${\mathbf H}$ is a diagonal matrix with $60\%$ diagonal elements randomly being zero, and the peak value of $\pmb x$ is set to be 2, 10, and 30, respectively. Restoration results are shown in Figure \ref{fig: result_Poisson_inpainting}. With the same percentage of missing pixels, the quality of the restored image and associated uncertainty map increases as the peak value increases. It is however worth noting that even when the peak value is very low (peak value = 2) and a large fraction of pixels is missing ($60\%$ missing), the proposed algorithm manages to restore a recognizable image and the uncertainty map also presents low uncertainty in the flat region (the body of \textit{Cameraman}). In addition, most pixels that are not in the 95$\%$ credible intervals construct the edges of objects. 

For these denoising tasks, VST-BM3D takes about 0.3 seconds, SPIRAL-TV takes about 1.7 seconds, PIDAL-TV takes about 1.5 seconds, and one expert of the proposed EP method takes about 513.3 seconds for 13 iteration (each iteration takes about 44 seconds). In Algorithm~\ref{alg: algorithm_Poisson_likeli}, the main cost of computation in each iteration includes: (\romannumeral 1) line 4, the 1D integrals in \eqref{Eq: moments_of_tilted_for_q1_u} for $n=1,\dots,256\times 256$ pixels in serial, (\romannumeral 2) lines 8-11, EP update of $q_{x,1}(\pmb x)$, and (\romannumeral 3) lines 16-19, EP update of $q_{x,0}(\pmb x)$. 

\subsubsection{Deconvolution}

\begin{figure}[!ht]
\centerline{\includegraphics[width=1\textwidth]{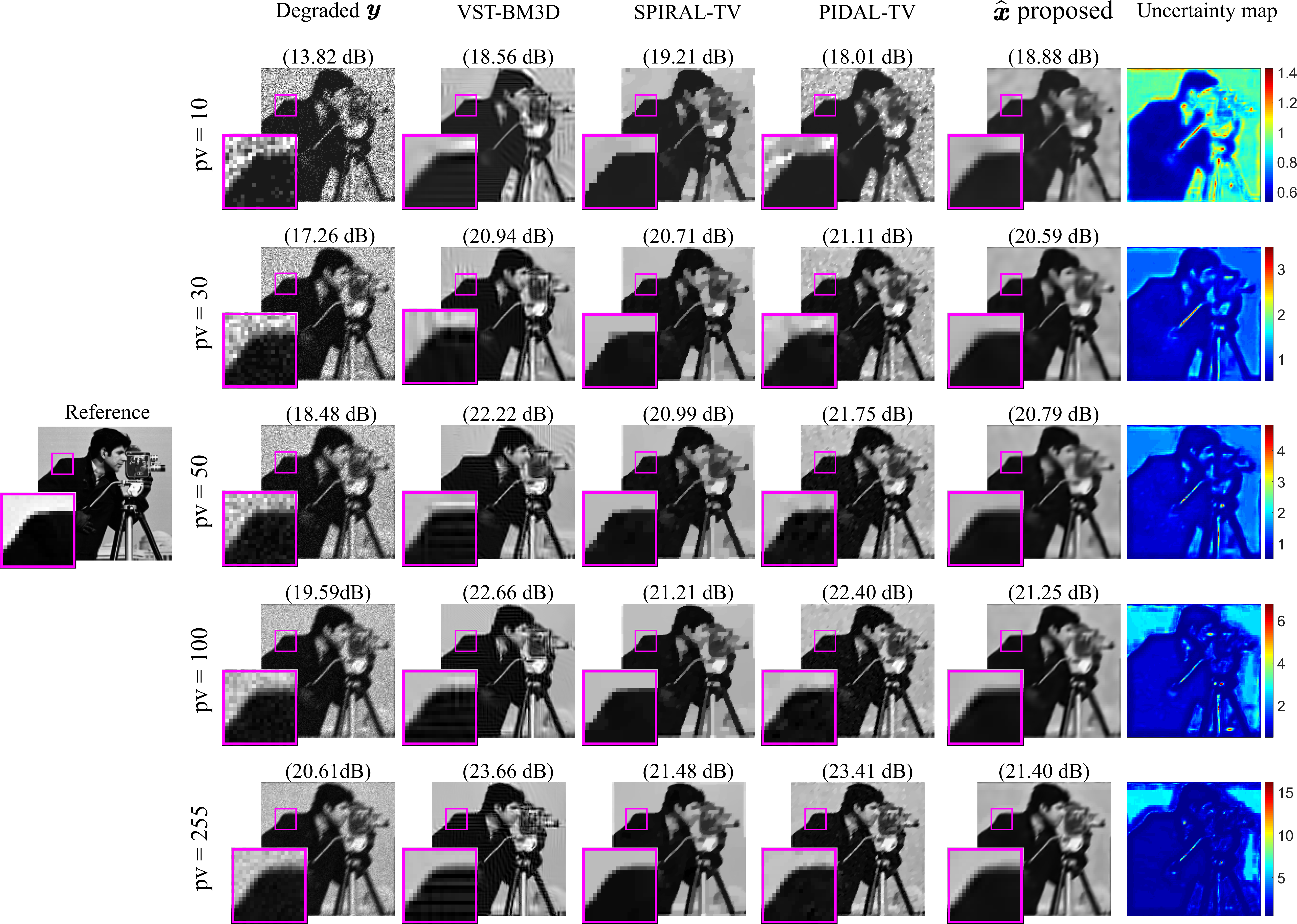}}
\caption{Results for deconvolution with Poisson noise. 
First column: reference of ground truth image and a zoomed-in region. Second column: observations degraded by a motion kernel of size $5\times 5$ pixels in 45 degree direction and Poisson noise with peak values $\{10,30,50,100,255\}$. Third column: estimated images by VST-BM3D \cite{makitalo2010optimal}. Fourth column: estimated images by SPIRAL-TV \cite{harmany2011spiral}. Fifth column: estimated images by PIDAL \cite{figueiredo2010restoration}. Sixth column:  estimated images by the proposed EP-EM method. Last column: uncertainty maps for the estimated images obtained by the proposed method.}
\label{fig: result_PoissonDeconv}
\end{figure}

The same $\textit{Cameraman}$ grayscale image as that used for deconvolution experiments with Gaussian noise is used here, but it is multiplied by peak values $\{10,30,50,100,255\}$ to obtain the ground truth $\pmb x$. The observed image $\pmb y$ is simulated by $\pmb y \sim {\mathcal P}({\mathbf H}\pmb x)$, and ${\mathbf H}$ is formed by a $5\times 5$ pixels motion kernel with an angle of 45 degrees. Figure \ref{fig: result_PoissonDeconv} presents examples of images restored by different methods. Although the proposed EP-EM method does not achieve the highest PSNRs, it does not produce artefacts as severe as VST-BM3D, SPIRAL-TV or PIDAL-TV. In addition, it can be seen that the overall uncertainty of the restored images increases as the peak value increases, which is due to the fact that as the mean of a Poisson distribution increases it induces larger posterior uncertainties, which are not completely reduced by the prior models. 

For deconvolution of images of size $128 \times 128$ pixels, VST-BM3D takes about 0.4 seconds, SPIRAL-TV takes about 3.7 seconds, PIDAL-TV takes about 13.8 seconds, and one expert of the proposed method takes about 869.3 seconds for 20 iterations (each iteration takes about 44  seconds on average. 20 samples are generated in RBMC to compute ${\rm Cov}_{P_1}(\pmb x)$ in \eqref{Eq: marginal_P_x}). The costly computation is similar to deconvolution with Gaussian i.i.d. noise, i.e., most of the computation can be reduced by parallel implementation.

In contrast to the Gaussian noise case, implementing more traditional EP algorithm is not straightforward with Poisson noise. Our method using auxiliary variables benefits from a cavity distribution with a diagonal covariance matrix when the tilted distribution is associated with the Poisson likelihood. This allows the parallel computation of the resulting marginal moments. Without auxiliary variables and assuming full covariance matrices for the approximating factors, the resulting EP updates would need to be performed sequentially, using rank-1 updates. In our experience, such updates can become numerically unstable, especially when enforcing the covariance matrices to be positive definite. Thus, we did not include comparisons with alternative EP implementations with less restrictive covariance structure constraints.

\section{Conclusions and discussions}
\label{Sec: Discussion and Conclusions}

In this paper, we proposed a new EP framework for image restoration using patch-based priors. In a similar fashion to VB, EP approximates the posterior distribution of interest by a simpler distribution whose moments are tractable. To address the computational challenges associated with large covariance matrix handling, we proposed to consider a combination of isotropic, diagonal and block-diagonal matrices, allowing, at least partially, distributed computation. Using augmented Bayesian models and factor graphs, we showed that the proposed method can be easily adapted to complex noise models. Indeed, for independently distributed noise realizations, the corresponding EP update simply requires the estimation of the first and second moments of 1D distributions, which can be performed in a parallel fashion. Moreover, if the initial GMM trained on external images whose scale does not match the current image of interest, it is possible to adjust some of its parameters (e.g., scale and offset) using a variational EM approach, which does not require crucial user supervision (in constrast to most MAP approaches). Although the proposed EP algorithm does not outperform state-of-the-art methods tailored for each restoration task in terms of quality of the estimated images, as an approximate MMSE estimator, it provides competitive posterior mean estimates with additional pixel-wise posterior uncertainty quantification, which holds great promises for uncertainty quantification in large scale imaging problems. Although we did not implement fully optimized versions of the EP-EM methods, the reported computational time can be further reduced by using a parallel update of blocks in covariance matrices of each expert, and parallel running for all the experts. This would however require significant additional work to optimize the updates for the different scenarios considered, and this is left to future work.

In practice, we observed that the Product-of-Experts approach provides very good results, and in particular, it allows blocky artefacts to be removed. Moreover, the resulting marginal uncertainties, are in practice good indicators of how far the posterior mean is from the actual pixel value. Hence, this information could be used for subsequent decision processes. It should however be noted that due to their approximate nature, such measures of uncertainty should be exploited with care. Since the exact posterior moments are however often not accessible, assessing the accuracy of the EP approximations remains an open problem.
 
Here, the block-diagonal covariance structure is motivated by the fact that the exact posterior covariance matrix is diagonally dominant. For deconvolution with Gaussian noise, we illustrated that the block-diagonal assumption does not significantly affect the estimated posterior means but it can modify the estimated marginal variances. Nonetheless, such structural constraints seemed necessary for EP to be scalable in the Poisson regime. If the blur kernel is larger than the patches considered, we observed that the reconstruction performance and convergence speed of EP can be degraded. Similarly, if the operator $\mathbf H$ introduces strong far-field correlation between pixels, as can happen in compressive sensing or tomography problems and when global image prior models are used, the method proposed in this paper does not provide satisfactory results. In such cases, different structural constraints should be investigated to make sure the EP method remains scalable and flexible enough to capture partially posterior correlations. While the EP covariance structures can be informed by the correlations expected to be present in the exact posterior distribution, they should also be designed and exploited together with the structures of the EP factor graphs, in particular when using data augmentation. For instance, even if $Q(\pmb x)$ is designed to have a block-diagonal covariance matrix, not all the approximating factors need to satisfy the same constraints.

\newpage

\bibliographystyle{IEEEtran}
\bibliography{biblio.bib}

\appendix
\section{Details of KL divergence minimization with difference covariance structures} 
\label{Appendix: Details of EP update for block-diagonal covariance matrices}
When minimizing the KL divergence in the form of $KL(P||Q)$, where $P$ is a tilted distribution and $Q$ is an EP Gaussian approximation, the divergence can be parameterized by the mean and covariance matrix of $Q$. Here for simplicity, $P$ and $Q$ can represent densities of $\pmb u$ or $\pmb x$. In the following derivation, we use the following general notation for the mean and covariance matrices, as listed in Table \ref{Table: Mean and covariance notations}. Moreover, we have ${\mathbf \Omega}_* = {\mathbf \Omega}_0 + {\mathbf \Omega}_1$, ${\mathbf \Omega}_*\pmb \mu_* = {\mathbf \Omega}_0\pmb \mu_0 + {\mathbf \Omega}_1 {\pmb \mu}_1$. 

\begin{table}[htbp] 
\label{Table: Mean and covariance notations}
\centering 
\begin{tabular}{l|c|ccc} 
\toprule                    
& $P(.)$ & $Q(.)$ & $q_0(.)$& $q_1(.)$  \\  
\hline  
Mean& ${\mathbb E}_P[.]$& $\pmb \mu_*$ & $\pmb \mu_0$ &$\pmb \mu_1$ \\
Covariance matrix & ${\rm Cov}_P(.)$& ${\mathbf \Sigma}_*$& ${\mathbf \Sigma}_0$ & ${\mathbf \Sigma}_1$\\
Precision matrix & & ${\mathbf \Omega}_* $& ${\mathbf \Omega}_0$& ${\mathbf \Omega}_1$\\
\bottomrule 
\end{tabular}
\caption{Notations for the mean and covariance matrices of $P$, $Q$, and individual approximating factors $q_0(.)$, $q_1(.)$.}
\end{table}
As explained in Section \ref{Sec: patch-basedEP_Gaussian_observation}, if the mean of $Q$ is unconstrained, the KL divergence is minimized when $\pmb \mu_* = {\mathbb E}_P[.]$, and we only need to optimize ${\mathbf \Omega}_*$. The loss function reduces to 
\begin{equation}
F({\mathbf \Omega}_i) \propto -\log(\det({\mathbf \Omega}_i+{\mathbf \Omega}_{\backslash i})) +\langle  {\mathbf \Omega}_i +{\mathbf \Omega}_{\backslash i}, {\rm Cov}_P(.)\rangle,
\label{Eq: F_Psi_i}
\end{equation}
which is convex w.r.t. ${\mathbf \Omega}_i \in \mathbb{S}_{++}^{\text{N}}$. Therefore, without additional constraints, ${\mathbf \Omega}_i$ can be obtained by solving 
\begin{equation}
\widehat {\bf \Omega}_i = \underset{{\mathbf \Omega}_i\in \mathbb{S}_{++}^{\text{N}}}{\textrm{argmin}}  -\log( {\rm det}({\mathbf \Omega}_i+{\mathbf \Omega}_{\backslash i})) +\langle {\bf \Omega}_i+{\mathbf \Omega}_{\backslash i}, {\rm Cov}_P(.)\rangle.
\end{equation}

\subsection{EP update of diagonal covariance}
\label{subsec: appen_diagonal covariance}
In Section \ref{Sec: patch-basedEP_Gaussian_observation}, we discussed the update of ${\mathbf \Omega}_{i}$, assuming that both ${\mathbf \Omega}_{i}$ and ${\mathbf \Omega}_{\backslash i}$ are block-diagonal. Here, we consider the case where both ${\mathbf \Omega}_{i}$ and ${\mathbf \Omega}_{\backslash i}$ are diagonal while optimizing ${\mathbf \Omega}_{i}$. For brevity, we use
$[d_{i,1},\dots,d_{i,N}] = {\textrm diag}({\rm Cov}_{P_i}(.))$, $[p_{i, 1},\dots,p_{i,N}]={\textrm diag}({\mathbf \Omega}_{i})$ and $[p_{\backslash i, 1},\dots,p_{\backslash i,N}]={\textrm diag}({\mathbf \Omega}_{\backslash i})$. In that case, \eqref{Eq: F_Psi_i} becomes
\begin{equation}
F({\bf \Omega}_i) \propto \sum\limits_{n=1}^{N}-\log(p_{i,n}+p_{\backslash i,n})+(p_{i,n}+p_{\backslash i,ii})d_{i,n},
\label{Eq: diagonal_v}
\end{equation}
which can be optimized independently w.r.t. the elements of ${\textrm diag}({\mathbf \Omega}_{i})$. For the element $p_{i,n}$, the cost function becomes
\begin{equation}
F(p_{i,n}) \propto -\log(p_{i,n}+p_{\backslash i,n})+(p_{i,n}+p_{\backslash i,n})d_{i,n}.
\end{equation}
This function is convex w.r.t. $p_{i,n}$ and has a unique minimizer, $p_{i,n} = \frac{1}{d_{i,n}} - p_{\backslash i,n}$. If this minimizer is negative, it is replaced by a small positive value (e.g., $10^{-8}$), leading to a large variance in ${\bf \Sigma}_i$ for that element. 

\subsection{EP update of isotropic covariance}
\label{subsec: appen_covariance_iso}

Based on \eqref{Eq: diagonal_v}, when the diagonal elements of the precision matrix ${\mathbf \Omega}_i$ are constrained to be equal to a same value denoted $p$, the loss function  becomes
\begin{equation}
F({\bf \Omega}_i) \propto \sum\limits_{n=1}^{N}-\log(p+p_{\backslash i,n})+(p+p_{\backslash i,n})d_{i,n}.
\end{equation}
This function is convex w.r.t. $p$ and it can be minimized using a Newton-Raphson method. At the $(t)$-th iteration, the update is 
\begin{equation}
    p_{(t)} = \max \left(p_{(t-1)}+ \frac{\sum\limits_{n=1}^N \frac{1}{p_{(t-1)} +p_{\backslash i, n}} -\sum\limits_{n=1}^N d_{i,n}}{\sum\limits_{n=1}^n \frac{1}{(p_{(t-1)}+p_{\backslash i,n})^2}}, \epsilon \right),
\end{equation}
where $\epsilon=10^{-8}$ is used to ensure that the covariance matrix remains positive definite.

\end{document}